\documentclass[conference]{IEEEtran}
\IEEEoverridecommandlockouts

\usepackage{cite}
\usepackage{amsmath,amssymb,amsfonts,amsbsy,amsthm}
\usepackage{graphicx}
\usepackage{textcomp}
\usepackage{xcolor}
\usepackage{hyperref}       
\usepackage{url}            
\usepackage{booktabs}       
\usepackage{amsfonts}       
\usepackage{amsmath}
\usepackage{nicefrac}       
\usepackage{microtype}      
\usepackage{graphicx}
\usepackage[skip=10pt,font=normalsize,labelfont=bf]{caption}
\usepackage{subcaption}
\usepackage{algorithm}
\usepackage{algpseudocode}
\usepackage{gensymb}
\usepackage{lipsum}
\usepackage{mathtools}
\usepackage{multirow}
\usepackage{subcaption}
\usepackage{pifont}
\usepackage{array}
\usepackage{textcase}
\usepackage{booktabs}
\usepackage{fancyhdr}
\usepackage{setspace}

\def\BibTeX{{\rm B\kern-.05em{\sc i\kern-.025em b}\kern-.08em
    T\kern-.1667em\lower.7ex\hbox{E}\kern-.125emX}}

\newtheorem{definition}{Definition}[section]

\newcommand{\nPr}[2]{\,^{#1}P_{#2}} 
\newcommand\Set[2]{\{\,#1\mid#2\,\}}

\newcommand{\norm}[1]{\left\lVert#1\right\rVert} 
\newcommand{\pluseq}{\mathrel{+}=}

\algnewcommand\algorithmicinput{\textbf{Input:}}
\algnewcommand\INPUT{\item[\algorithmicinput]}
\algnewcommand\algorithmicoutput{\textbf{Output:}}
\algnewcommand\OUTPUT{\item[\algorithmicoutput]}
\algnewcommand\algorithmicparameters{\textbf{Parameters:}}
\algnewcommand\PARAMETERS{\item[\algorithmicparameters]}
\algnewcommand\algorithmicinitialize{\textbf{Initialize:}}
\algnewcommand\INITIALIZE{\item[\algorithmicinitialize]}
\algtext*{EndWhile}
\algtext*{EndIf}
\algtext*{EndFor}
\algtext*{EndFunction}
\algtext*{EndProcedure}
\algrenewcommand\textproc{\texttt}

\newcolumntype{U}{>{\centering\arraybackslash}p{2.9cm}}
\newcolumntype{V}{>{\centering\arraybackslash}p{2.0cm}}
\newcolumntype{T}{>{\centering\arraybackslash}p{1.5cm}}
\newcolumntype{X}{>{\centering\arraybackslash}p{1.65cm}}
\newcolumntype{Y}{>{\centering\arraybackslash}p{0.25cm}}
\newcolumntype{Z}{>{\centering\arraybackslash}p{0.55cm}}
\newcolumntype{P}{>{\centering\arraybackslash}p{1.95cm}}
\newcolumntype{S}{>{\centering\arraybackslash}p{0.65cm}}

\fancypagestyle{FirstPage}{
    \lfoot{
        \begin{spacing}{0.7}
            {\footnotesize This is an extended version of the paper (DOI: 10.1109/MDM55031.2022.00028) that was accepted for presentation at the 23rd IEEE International Conference on Mobile Data Management (MDM) 2022, Paphos, Cyprus. \textbf{IEEE Copyright Notice:} \copyright2022 IEEE. Personal use of this material is permitted.  Permission from IEEE must be obtained for all other uses, in any current or future media, including reprinting/republishing this material for advertising or promotional purposes, creating new collective works, for resale or redistribution to servers or lists, or reuse of any copyrighted component of this work in other works.}
        \end{spacing}
    } 
}
\begin{document}

\title{Reachability Embeddings: Scalable Self-Supervised Representation Learning from Mobility Trajectories for Multimodal Geospatial Computer Vision}

\author{
    \IEEEauthorblockN{
        Swetava Ganguli*\thanks{*Corresponding author. Alternative email: swetava@cs.stanford.edu.},
        C. V. Krishnakumar Iyer, 
        Vipul Pandey}
    \IEEEauthorblockA{
        Apple, Cupertino, CA \\
        \texttt{\{swetava,cvk,vipul\}@apple.com}
    }
}

\maketitle

\begin{abstract}
Self-supervised representation learning techniques utilize large datasets without semantic annotations to learn meaningful, universal features that can be conveniently transferred to solve a wide variety of downstream supervised tasks. In this paper, we propose a self-supervised method for learning representations of geographic locations from unlabeled GPS trajectories to solve downstream geospatial computer vision tasks. Tiles resulting from a raster representation of the earth's surface are modeled as nodes on a graph or pixels of an image. GPS trajectories are modeled as allowed Markovian paths on these nodes. A scalable and distributed algorithm is presented to compute image-like representations, called \textit{reachability summaries}, of the spatial connectivity patterns between tiles and their neighbors implied by the observed Markovian paths. A convolutional, contractive autoencoder is trained to learn compressed representations, called \textit{reachability embeddings}, of reachability summaries for every tile. Reachability embeddings serve as task-agnostic, feature representations of geographic locations. Using reachability embeddings as pixel representations for five different downstream geospatial tasks, cast as supervised semantic segmentation problems, we quantitatively demonstrate that reachability embeddings are semantically meaningful representations and result in 4--23\% gain in performance, as measured using area under the precision-recall curve (AUPRC) metric, when compared to baseline models that use pixel representations that do not account for the spatial connectivity between tiles. Reachability embeddings transform sequential, spatiotemporal mobility data into semantically meaningful tensor representations that can be combined with other sources of imagery and are designed to facilitate multimodal learning in geospatial computer vision.
\end{abstract}

\begin{IEEEkeywords}
Self-Supervised Learning, Representation Learning, Markov Chains, Multimodal Machine Learning
\end{IEEEkeywords}

\section{Introduction}\label{introduction}
\thispagestyle{FirstPage}

\subsection{Background}\label{introduction_background}
Graphs are natural data structures for applications in diverse domains \cite{leskovec20} e.g., recommendation systems, communication, social, and biological networks. Geospatial datasets (e.g., road networks, point clouds, 3D object meshes) can organically be represented as graphs with natural definitions of nodes and edges. Machine learning algorithms for graph analysis require feature vector representations of nodes, edges, substructures, or the whole graph. Instead of hand-engineering task-specific and domain-specific features, recent methods \cite{cai18} have focused on automatically learning low-dimensional, feature vector representations of graphs (\textit{graph embeddings}) and their components (e.g., \textit{node embeddings}). Reachability of one node from another is a fundamental concept in graph theory.

In parallel, \textit{self-supervised learning} (SSL) has been an area of active research and has achieved promising performance on both natural language processing (NLP) \cite{brown2020gpt3,devlin2018bert,mikolov13} and computer vision tasks \cite{chen2020simclr,grill20}. Bypassing the need for large, clean, labeled datasets which are expensive to produce in time and money, SSL often uses predefined \textit{pretext tasks} to derive supervision signals directly from unlabeled data by making neural networks predict withheld parts or properties of inputs. SSL aims to learn semantically meaningful, task-agnostic representations of data that can be used as inputs by downstream (usually supervised) task-specific models. SSL has been used to learn context-independent \cite{mikolov13} and contextual \cite{brown2020gpt3,devlin2018bert}, task-agnostic word embeddings for NLP applications. Most popular SSL techniques for learning visual representations can be classified into two types: \textit{generative} approaches that learn representations under the \textit{pretext} of generating images by modeling the data distribution \cite{donahue2019large,kingma2013vae,goodfellow2014gan}, and \textit{discriminative} approaches that use \textit{pretext tasks}, designed to efficiently produce labels for inputs (e.g., based on heuristics \cite{doersch15,gidaris18,noroozi16} or contrastive learning \cite{chen2020simclr,grill20}), coupled with a supervised objective.

Geospatial computer vision research has largely focused on active (e.g., SAR) and passive (e.g., optical) imagery data \cite{ganguli19,ma19,perez19}. This paper proposes a novel technique to make GPS trajectories, which are sequential, spatiotemporal datasets with rich geospatial connectivity information, amenable to computer vision-based analysis by itself or in combination with other data modalities in a multimodal setting. GPS and mobility datasets have been used for diverse applications \cite{wang20} including transportation modeling \cite{zheng15}, public safety and health policy \cite{smith20}, and building mapping services \cite{trinity21,xiao20}.

\subsection{Present Work and Contributions}\label{introduction_present_work}
We present a novel computer vision-based, self-supervised method for learning task-agnostic feature representations of geographic locations from GPS trajectories by modeling tiles resulting from a raster representation of the earth's surface as nodes of a graph (termed \textit{earth surface graph} (ESG)) and modeling the \textit{observed} GPS trajectories as evidence of \textit{allowed} Markovian paths on this graph. Equivalently, the proposed method learns node embeddings, for all nodes in the ESG, from observed paths on the ESG. The proposed method has two stages. For each node in the ESG, the first stage (Sections \ref{methodology_reachability_embeddings}--\ref{scalability_of_algorithm}) generates an \textit{image-like} representation, termed \textit{reachability summary}, of the connectivity pattern of the node with its neighbors inferred using observed trajectories passing through the node during a predefined observation time interval, $\Delta t$. A scalable and distributed algorithm is proposed for the first stage. The second stage (Section \ref{encoding_reachability_summaries}) generates $\boldsymbol{d}_{R}$-dimensional, compressed representations from reachability summaries, called \textit{reachability embeddings}, using a fully-convolutional, contractive autoencoder. In Appendix Section \ref{methodology_chapman_kolmogorov}, reachability embeddings are theoretically motivated and interpreted using the Chapman-Kolmogorov Equations (CKE) for Markov chains.

Experiments in Section \ref{experiments_and_results} evaluate the impact of reachability embeddings on solving downstream geospatial computer vision tasks using supervised semantic segmentation. The model inputs are combinations of image-like derivatives of road network and GPS trajectory data (Section \ref{dataset_description}) with pixels corresponding to nodes (convenient by design) of the ESG. For five different downstream tasks, we show that using reachability embeddings as input features outperforms baselines (models with input features that do not account for spatial connectivity between nodes) resulting in 4--23\% gain in AUPRC (area under precision-recall curve) while requiring upto 67\% less trajectory data than the baselines. Thus, reachability embeddings are semantically meaningful representations of geographic locations and denser, more informative representations derived from trajectories compared to those not accounting for spatial connectivity between nodes. Experiments in the latter part of Section \ref{experiments_and_results} demonstrate that representing mobility data as reachability embeddings facilitates multimodal modeling using computer vision techniques. Pixel-wise embeddings yield multi-channel image-like tensors. Thus, by design, pixel-wise alignment of data modalities implies spatial alignment and early fusion is feasible via concatenation of image-like tensor representations of various data modalities along the channel dimension.

Presence of geospatial features (e.g., traffic control devices, physical traffic restrictions, traffic infrastructure) at a location, $\boldsymbol{\ell}$, imposes constraints on traffic flow that manifests in the (i) frequency of, (ii) directionality (from or to $\boldsymbol{\ell}$) of, (iii) time taken during, and (iv) distance traveled during observed transitions between $\boldsymbol{\ell}$ and its neighboring locations across various motion modalities (e.g., walking, driving, biking). Reachability summaries, and consequently reachability embeddings, are designed to be \textit{contextual} representations (analogous to advantage of contextual \cite{devlin2018bert} over fixed \cite{mikolov13} word embeddings in NLP) that encode these motion characteristics and traffic flow patterns occurring during $\Delta t$. Events that significantly change traffic patterns like road construction, road closures, seasonal effects (e.g., winter vs. summer), etc. change the computed embeddings making them a good input feature candidate for geospatial feature or change detection models. 

The following are the main contributions of this paper: (1) A novel, two-stage, computer vision-based, self-supervised procedure to learn low dimensional, geospatial feature representations, called reachability embeddings, from GPS trajectories is proposed; (2) A scalable and distributed algorithm that can incorporate distance traveled and time taken during a trajectory for computing these embeddings is presented; (3) Enhanced performance on five downstream geospatial feature detection tasks and the ease of multimodal modeling with reachability emebeddings is demonstrated; (4) A theoretical interpretation of reachability embeddings using CKE is provided.

While the presentation in this paper is targeted towards geospatial and remote sensing applications, the concept of generating node embeddings based on reachability can be extended to graphs in other domains. For example, (i) graph traversal algorithms can be used to assemble node reachability summaries; (ii) sampled random walk paths, used in \cite{dong17,grover16,perozzi14}, can be used analogous to trajectories to compute reachability summaries for nodes. The generated summaries can be subsequently encoded into reachability embeddings.

\setlength{\belowcaptionskip}{-2pt}
\setlength{\textfloatsep}{5pt}
\begin{figure*}
    \centering
    \includegraphics[width=0.90\textwidth]{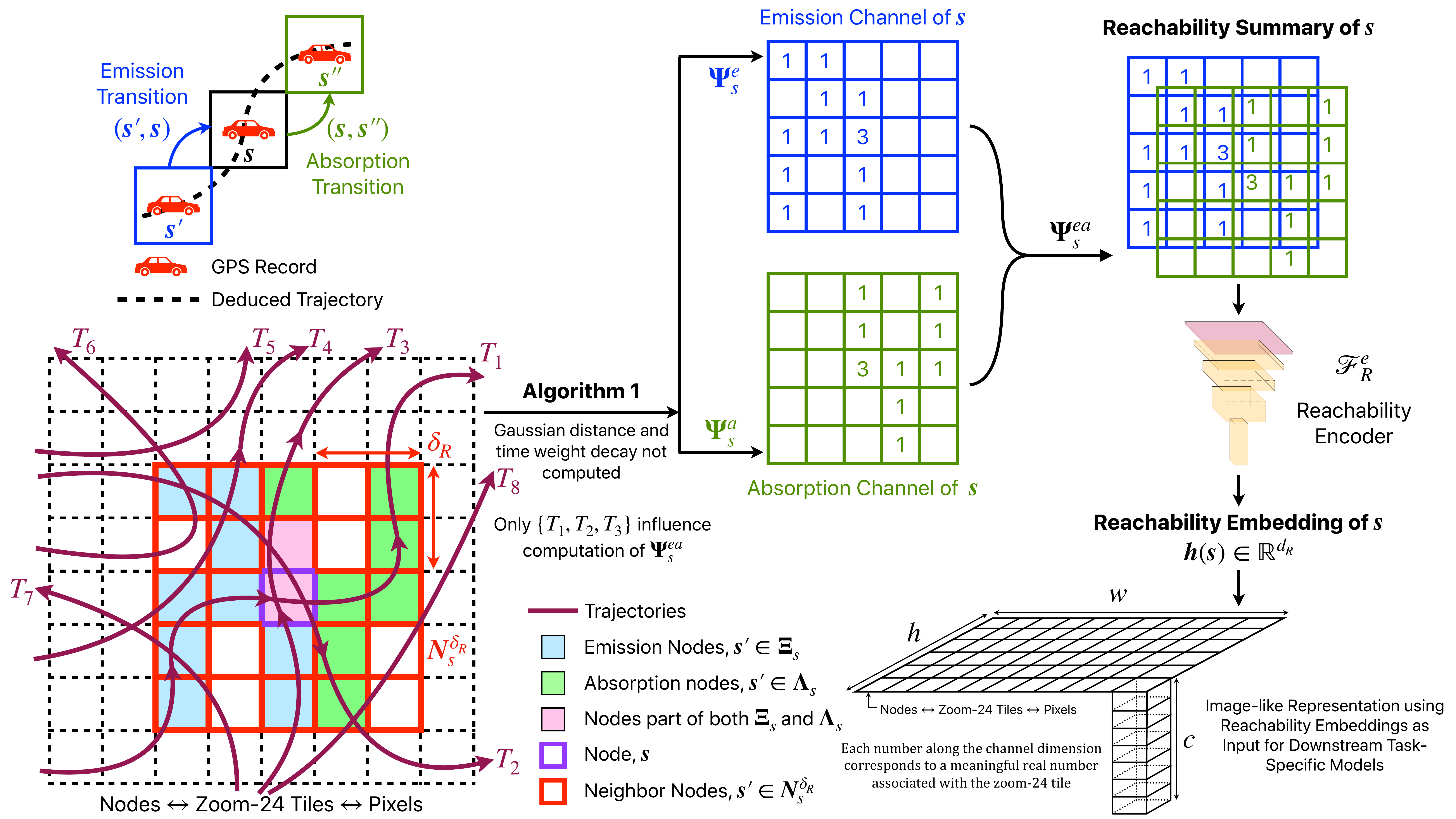}
    \caption{Schematic describing the algorithm for generation of reachability embeddings from mobility trajectories.}
    \label{image_like_representation}
\end{figure*}

\section{Notation and Preliminaries}\label{notation_and_preliminaries}
Figure \ref{image_like_representation} provides a reference schematic for notation and terminology introduced in this and the next section.
\begin{definition}\label{def_zoom_24_tile}
    A \textbf{zoom-q tile} is the cell or tile resulting from a $2^{q} \times 2^{q}$ raster representation of the spherical Mercator projection of the earth's surface \cite{epsg_3857}. With the northwest corner tile as origin, every tile, $\boldsymbol{s}$, is assigned 2D, non-negative integer coordinates, $(\boldsymbol{x}_s, \boldsymbol{y}_s)$, which increase in the $\boldsymbol{x}$-direction towards the east (right) and in the $\boldsymbol{y}$-direction towards the south (down).
\end{definition}
\begin{definition}\label{def_gps_trajectory}
    A GPS trajectory, $\boldsymbol{T}_i \in \boldsymbol{\mathcal{T}}(t_0, \Delta t)$, or simply \textbf{trajectory}, is a discrete, sequential, and chronological representation of the spatiotemporal movement of an object \cite{zheng15} consisting of sequence of ordered pairs, $\boldsymbol{T}_i = (\boldsymbol{p}_1^i, \ldots, \boldsymbol{p}_{n_i}^i)$, where pair, $\boldsymbol{p}^i_k = (\boldsymbol{z}_k^i, \boldsymbol{t}_k^i)$, for $k = 1, \ldots, n_i$. $\boldsymbol{p}^i_k$ is called a GPS record or simply a \textbf{record}. $\boldsymbol{\mathcal{T}}(t_0,\Delta t)$ is the set of $|\boldsymbol{\mathcal{T}}(t_0, \Delta t)| = M$ trajectories occurring within observation time interval $\Delta t$ starting at $t_0$, $n_i$ is length of trajectory $\boldsymbol{T}_i$, and $\boldsymbol{z}^i_k$ is the zoom-24 tile\footnote{Corresponds to spatial resolution $\approx 2.38$m at the equator. Finite area tiles mitigate handling real-valued latitude-longitude GPS location pairs. The tile represents all such pairs within itself.} where the object with trajectory $\boldsymbol{T}_i$ was present at time $\boldsymbol{t}_k^i$. Note: $\boldsymbol{t}_1^i < \ldots < \boldsymbol{t}_{n_i}^i$ for $i = 1, \ldots, M$.
\end{definition}
\begin{definition}\label{def_implied_allowed_transition_set}
    The \textbf{implied allowed transitions multiset}, $\boldsymbol{A}_{i}^m$, associated with trajectory $\boldsymbol{T}_i$ is the unordered multiset of all $\nPr{n_i}{2} = n_i(n_i-1)$ ordered pairs of zoom-24 tiles, $(\boldsymbol{z}_k,\boldsymbol{z}_l)$, that can be constructed from trajectory $\boldsymbol{T}_i$ such that $l \geq k$ . The unordered set consisting of only the unique ordered pairs is called the \textbf{implied allowed transitions set}, $\boldsymbol{A}_{i}^s$. 
\end{definition}
Ordered pairs generated from the trajectories are modeled as Markovian state transitions over the countably finite state space of zoom-24 tiles. Hence, the trajectory index is dropped from the the ordered pairs in def. (\ref{def_implied_allowed_transition_set}).
\begin{definition}\label{def_earth_surface_graph}
    The \textbf{earth surface graph} (ESG), $\boldsymbol{G}_{ES}(\boldsymbol{V}_{ES},\boldsymbol{E}_{ES})$, is an inferred, weighted directed graph representation of the earth's surface where zoom-24 tiles form the nodes, $\boldsymbol{V}_{ES}$, and the set union of $\boldsymbol{A}_{i}^s$ of all $\boldsymbol{T}_i \in \boldsymbol{\mathcal{T}}(t_0,\Delta t)$ forms the edge set, $\boldsymbol{E}_{ES}$. For nodes, $\boldsymbol{s},\boldsymbol{s}' \in \boldsymbol{V}_{ES}$, $\boldsymbol{c}_{i}^{(s,s')}$ is the number of occurrences of the transition $(\boldsymbol{s},\boldsymbol{s}')$ in $\boldsymbol{A}_{i}^m$. The directed edge, $(\boldsymbol{s},\boldsymbol{s}')$, in $\boldsymbol{E}_{ES}$ is assigned the weight, $\boldsymbol{w}_{ES}^{(s,s')} = \sum\limits_{i=1}^{i=M} \boldsymbol{c}_{i}^{(s,s')}$.
\end{definition}
Definitions (\ref{def_gps_trajectory}), (\ref{def_implied_allowed_transition_set}), and (\ref{def_earth_surface_graph}) present the equivalences used in this paper: (i) zoom-24 tiles $\leftrightarrow$ Markovian state space $\boldsymbol{V}_{ES}$ $\leftrightarrow$ nodes of $\boldsymbol{G}_{ES}$, (ii) GPS trajectory $\leftrightarrow$ Markovian trajectory in $\boldsymbol{V}_{ES}$ $\leftrightarrow$ path on $\boldsymbol{G}_{ES}$. 
\begin{definition}\label{def_reachability}
    Given two nodes, $\boldsymbol{s}_1, \boldsymbol{s}_2 \in \boldsymbol{V}_{ES}$, $\boldsymbol{s}_2$ is \textbf{reachable} from $\boldsymbol{s}_1$ if $\,\,\exists\,\, (\boldsymbol{s}_1,\boldsymbol{s}_2) \in \boldsymbol{E}_{ES}$. 
\end{definition}
While $q=24$ in this paper, the algorithms presented in Section \ref{methodology}, are generic and treat $q$ as a parameter.

\section{Methodology}\label{methodology}

\subsection{Reachability Embeddings}\label{methodology_reachability_embeddings}
The proposed method learns a mapping function, $\boldsymbol{f}_R: \boldsymbol{V}_{ES} \rightarrow \mathbb{R}^{d_R}$, from nodes in the ESG, $\boldsymbol{V}_{ES}$, to real valued, $\boldsymbol{d}_{R}$-dimensional feature representations we call reachability embeddings. Each record in a trajectory is associated with a motion modality (e.g., driving, walking, biking). A pre-trained, LSTM-based, neural motion modality filter is used to keep only the records corresponding to a chosen modality. Henceforth, we will assume that all records in all trajectories in $\boldsymbol{\mathcal{T}}(t_0,\Delta t)$ have the same motion modality.

Given a node, $\boldsymbol{s} \in \boldsymbol{V}_{ES}$, two sets of nodes can naturally be defined relative to $\boldsymbol{s}$ based on observed reachability patterns in trajectories in $\boldsymbol{\mathcal{T}}(t_0,\Delta t)$. The first set, called the \textit{emission set} for node $\boldsymbol{s}$, $\boldsymbol{\Xi}_{s}$, consists of nodes \textit{from which} transitions to $\boldsymbol{s}$ are observed. Such transitions are termed \textit{emission transitions}. The second set, called the \textit{absorption set} for node $\boldsymbol{s}$, $\boldsymbol{\Lambda}_{s}$, consists of nodes \textit{to which} transitions from $\boldsymbol{s}$ are observed. Such transitions are termed \textit{absorption transitions}. Thus,
\begin{subequations}
    \begin{gather}
        \boldsymbol{\Xi}_s  \coloneqq \Set{\boldsymbol{s}'}{\boldsymbol{s}' \in \boldsymbol{V}_{ES}\,\,\text{and}\,\,(\boldsymbol{s}',\boldsymbol{s}) \in \boldsymbol{E}_{ES}},\label{emission_set}\\
        \boldsymbol{\Lambda}_s \coloneqq \Set{\boldsymbol{s}'}{\boldsymbol{s}' \in \boldsymbol{V}_{ES}\,\,\text{and}\,\,(\boldsymbol{s},\boldsymbol{s}') \in \boldsymbol{E}_{ES}}.\label{absorption_set}
    \end{gather}
\end{subequations}
Owing to transitions from $\boldsymbol{s}$ to itself, $\boldsymbol{s} \in \boldsymbol{\Xi}_s, \boldsymbol{\Lambda}_s$. With the intuition that transitions occurring over large spatial distances should not influence the learned representations of nodes, transitions occurring only within a specified buffer, termed the \textit{reachability buffer}, $\boldsymbol{\delta}_R \in \mathbb{Z}^{+}$, are considered valid. The set of nodes within $\boldsymbol{\delta}_R$ of node $\boldsymbol{s}$, called the \textit{reachable neighborhood}, $\boldsymbol{N}_{s}^{\delta_R}$, are 
\begin{equation}\label{reachability_neighborhood}
    \boldsymbol{N}_{s}^{\delta_R} \coloneqq \Set{\boldsymbol{s}'}{\boldsymbol{s}' \in \boldsymbol{V}_{ES}\text{,} |\boldsymbol{x}_s - \boldsymbol{x}_{s'}| \leq \boldsymbol{\delta}_R\text{,} |\boldsymbol{y}_s - \boldsymbol{y}_{s'}| \leq \boldsymbol{\delta}_R}.
\end{equation}
Combining definitions (\ref{emission_set}), (\ref{absorption_set}), and (\ref{reachability_neighborhood}), the set of valid emission and absorption transitions with respect to node $\boldsymbol{s}$, denoted as $\boldsymbol{\Xi}_{s}^{\delta_R}$ and $\boldsymbol{\Lambda}_{s}^{\delta_R}$ respectively, are
\begin{equation}
    \boldsymbol{\Xi}_{s}^{\delta_R} = \boldsymbol{\Xi}_{s} \cap \boldsymbol{N}_{s}^{\delta_R}\;\;\;\;\;\text{and}\;\;\;\;\;\boldsymbol{\Lambda}_{s}^{\delta_R} = \boldsymbol{\Lambda}_{s} \cap \boldsymbol{N}_{s}^{\delta_R}.
\end{equation}
The Cartesian grid structure of nodes in $\boldsymbol{V}_{ES}$ can be exploited to create an \textit{image-like} representation of the reachability pattern for all nodes, $\boldsymbol{s}$, implied by their $\boldsymbol{\Xi}_{s}^{\delta_R}$ and $\boldsymbol{\Lambda}_{s}^{\delta_R}$. With $\boldsymbol{L} = 2\boldsymbol{\delta}_R + 1$, for each $\boldsymbol{s} \in \boldsymbol{V}_{ES}$, construct two zero-initialized $\boldsymbol{L} \times \boldsymbol{L}$ square matrices, $\boldsymbol{\Psi}_{s}^{e}$ (\textit{emission channel}) and $\boldsymbol{\Psi}_{s}^{a}$ (\textit{absorption channel}), with the top-left corner entries assigned the index $(0,0)$, first index increasing rightward, and second index increasing downward. For every $\boldsymbol{s}' \in \boldsymbol{\Xi}_{s}^{\delta_R}$ and $\boldsymbol{s}'' \in \boldsymbol{\Lambda}_{s}^{\delta_R}$, set
\begin{subequations}
    \begin{gather}
        \boldsymbol{\Psi}_{s}^{e}[\boldsymbol{x}_{s'} - \boldsymbol{x}_{s} + \boldsymbol{\delta}_R, \boldsymbol{y}_{s'} - \boldsymbol{y}_{s} + \boldsymbol{\delta}_R] = \boldsymbol{w}_{ES}^{(s',s)},\label{emission_matrix}\\
        \boldsymbol{\Psi}_{s}^{a}[\boldsymbol{x}_{s''} - \boldsymbol{x}_{s} + \boldsymbol{\delta}_R, \boldsymbol{y}_{s''} - \boldsymbol{y}_{s} + \boldsymbol{\delta}_R] = \boldsymbol{w}_{ES}^{(s,s'')}.\label{absorption_matrix}
    \end{gather}
\end{subequations}
The $(\boldsymbol{\delta}_R, \boldsymbol{\delta}_R)$-indexed entry equals $\boldsymbol{w}_{ES}^{(s,s)}$ in both, $\boldsymbol{\Psi}_{s}^{e}$ and $\boldsymbol{\Psi}_{s}^{a}$. Matrices $\boldsymbol{\Psi}_{s}^{e}$ and $\boldsymbol{\Psi}_{s}^{a}$, have three important properties: (i) they preserve the relative spatial proximity of nodes in $\boldsymbol{N}_{s}^{\delta_R}$ with respect to $\boldsymbol{s}$; (ii) $\boldsymbol{\Psi}_{s}^{e}$ and $\boldsymbol{\Psi}_{s}^{a}$ are the spatial-relation preserving matrix representations of the column and row corresponding to $\boldsymbol{s}$ of the adjacency matrix of nodes in $\boldsymbol{N}_{s}^{\delta_R}$; (iii) normalization by matrix sums yields the Markovian transition probabilities to $\boldsymbol{s}$ from all $\boldsymbol{s}' \in \boldsymbol{N}_{s}^{\delta_R}$ and from $\boldsymbol{s}$ to all $\boldsymbol{s}'' \in \boldsymbol{N}_{s}^{\delta_R}$, respectively. Matrices, $\boldsymbol{\Psi}_{s}^{e}$ and $\boldsymbol{\Psi}_{s}^{a}$, are treated as two channels and stacked along the channel dimension to create a $\boldsymbol{L} \times \boldsymbol{L} \times 2$-dimensional, image-like entity, called the \textit{reachability summary}, $\boldsymbol{\Psi}_{s}^{ea}$, for node $s$. A contractive \cite{rifai11}, fully-convolutional autoencoder, $\mathcal{F}_R$, is trained to generate compressed, robust, low-dimensional representations of $\boldsymbol{\Psi}_{s}^{ea}$, called the \textit{reachability embedding}, $\boldsymbol{h}(\boldsymbol{s}) \in \mathbb{R}^{d_R}$, for node $\boldsymbol{s}$ using contractive reconstruction of $\boldsymbol{\Psi}_{s}^{ea}$ as the self-supervision task for representation learning. Reachability embeddings, produced for every node $\boldsymbol{s} \in \boldsymbol{V}_{ES}$ by employing the encoder component of $\mathcal{F}_R$, captures the spatial connectivity of node $\boldsymbol{s}$ to nodes in $\boldsymbol{N}_{s}^{\delta_R}$ as evidenced by observed trajectories (modeled as Markovian) in $\mathcal{T}(t_0,\Delta t)$.

\subsection{Distributed data-parallel algorithm for generating reachability summaries}\label{data_parallel_distributed_algorithm}
Computing $\boldsymbol{\Psi}_{s}^{ea}$ for all $\boldsymbol{s} \in \boldsymbol{V}_{ES}$ requires analysing all $M$ trajectories in $\mathcal{T}(t_0, \Delta t)$ and computing upto $2^{2q}$ matrices of size $\boldsymbol{L} \times \boldsymbol{L} \times 2$, where $q=24$. In practice, however, given a choice of motion modality, only a tractable subset (e.g., no driving or walking on water bodies) of the zoom-24 tiles (called \textit{active tiles} and denoted as $\overline{\boldsymbol{V}}_{ES}$) have trajectories that pass through them. The embedding vector is set to zero for non-active tiles. Algorithm \ref{algorithm_reachability_summary} proposes an efficient method to compute $\boldsymbol{\Psi}_{s}^{ea}\,\forall\,\boldsymbol{s} \in \overline{\boldsymbol{V}}_{ES}$ from $\mathcal{T}(t_0, \Delta t)$ assuming all its trajectories are of the same, user-chosen motion modality. The key idea in Algorithm \ref{algorithm_reachability_summary} is to compute $\boldsymbol{w}_{ES}^{(\boldsymbol{s}_1,\boldsymbol{s}_2)}, \boldsymbol{s}_1, \boldsymbol{s}_2 \in \overline{\boldsymbol{V}}_{ES}$ for all unique $(\boldsymbol{s}_1, \boldsymbol{s}_2)$ pairs (emission and absorption transitions) independently in parallel and then assemble the dense tensor, $\boldsymbol{\Psi}_{s}^{ea}$, for each $\boldsymbol{s} \in \overline{\boldsymbol{V}}_{ES}$. For node $\boldsymbol{s}$, let $\boldsymbol{S}_{s}^{a}: r_{s}(\boldsymbol{s}') \rightarrow c_{s}^{a}(\boldsymbol{s}')$ be a map constructed from observed absorption transitions $(\boldsymbol{s}, \boldsymbol{s}')$ for $\boldsymbol{s}' \in \boldsymbol{N}_{s}^{\delta_R}$. Here, $r_{s}(\boldsymbol{s}')$ is the row-major index of $\boldsymbol{s}'$ in $\boldsymbol{\Psi}_{s}^{a}$ and $c_{s}^{a}(\boldsymbol{s}')$ is the frequency of observing the transition $(\boldsymbol{s},\boldsymbol{s}')$, which equals $\boldsymbol{w}_{ES}^{(s,s')}$ when Algorithm \ref{algorithm_reachability_summary} finishes. Analogously, define $\boldsymbol{S}_{s}^{e}: r_{s}(\boldsymbol{s}') \rightarrow c_{s}^{e}(\boldsymbol{s}')$ with $c_{s}^{e}(\boldsymbol{s}')$ being frequency of observing transition $(\boldsymbol{s}',\boldsymbol{s})$ which equals $\boldsymbol{w}_{ES}^{(s',s)}$ when algorithm \ref{algorithm_reachability_summary} finishes. The reachability map, $\boldsymbol{S}: \boldsymbol{s} \rightarrow (\boldsymbol{S}_{s}^{e}, \boldsymbol{S}_{s}^{a})$, is a map from each node $\boldsymbol{s} \in \overline{\boldsymbol{V}}_{ES}$ to the tuple of the node's emission and absorption transition maps, $\boldsymbol{S}_{s}^{e}$ and $\boldsymbol{S}_{s}^{a}$. $\boldsymbol{S}_{s}^{e}$ and $\boldsymbol{S}_{s}^{a}$ are sparse representations of the information required to assemble $\boldsymbol{\Psi}_{s}^{e}$ and $\boldsymbol{\Psi}_{s}^{a}$ (i.e., $\boldsymbol{\Psi}_{s}^{ea}$).

\subsection[Capturing more than spatial connectivity]{Capturing more than spatial connectivity in $\boldsymbol{\Psi}_s^e, \boldsymbol{\Psi}_s^a, \boldsymbol{\Psi}_s^{ea}$}\label{encoding_more_information_in_reachability_summaries}
The emission ($\boldsymbol{\Psi}_s^e$) and absorption ($\boldsymbol{\Psi}_s^a$) channels, as defined in Section \ref{methodology_reachability_embeddings}, capture only the spatial connectivity between two nodes. In Algorithm \ref{algorithm_reachability_summary}, if the transition $(\boldsymbol{s},\boldsymbol{s}')$, with $\boldsymbol{s}' \in \boldsymbol{N}_{s}^{\delta_R}$, is observed in trajectory $\boldsymbol{T}_i$, $\boldsymbol{\Psi}_{s}^{a}(\boldsymbol{s'})$ and $\boldsymbol{\Psi}_{s'}^{e}(\boldsymbol{s})$ are both incremented by 1. These increments, denoted as $\Delta\boldsymbol{\Psi}_{s}^{a}(\boldsymbol{s}')$ and $\Delta\boldsymbol{\Psi}_{s'}^{e}(\boldsymbol{s})$, are independent of the distance covered, $\Delta_{d,i}^{(s,s')}$, or time taken, $\Delta_{t,i}^{(s,s')} = \boldsymbol{t}_{s'}^i - \boldsymbol{t}_{s}^i$, to reach $\boldsymbol{s}'$ from $\boldsymbol{s}$ in trajectory $\boldsymbol{T}_i$. Note that time taken and distance covered depend on the trajectory if multiple paths between the two nodes exist. Gaussian weight decay is used to incorporate this information into $\boldsymbol{\Psi}_{s}^{a}(\boldsymbol{s'})$ and $\boldsymbol{\Psi}_{s'}^{e}(\boldsymbol{s})$. The Gaussian with parameters $\mu$ and $\sigma$ is $G(\mu,\sigma)=(1/\sqrt{2\pi}\sigma)e^{-\mu^2/2\sigma^2}$. Given $\sigma_d$ and $\sigma_t$, two user-supplied hyperparameters that control the strengths of the distance and time weight decays, respectively, the weighted increment\footnote{The weight-decays can be thought of as the \textit{attention} payed by $\boldsymbol{s}$ to a neighboring node $\boldsymbol{s}' \in \boldsymbol{N}_{s}^{\delta_R}$ analogous to the soft-attention mechanism.} in $\boldsymbol{\Psi}_{s}^{a}(\boldsymbol{s'})$ and $\boldsymbol{\Psi}_{s'}^{e}(\boldsymbol{s})$ associated with the transition $(\boldsymbol{s},\boldsymbol{s'})$ in trajectory $\boldsymbol{T}_i$, is $\Delta\boldsymbol{\Psi}_s^a(\boldsymbol{s}') = \Delta\boldsymbol{\Psi}_{s'}^e(\boldsymbol{s}) \coloneqq \Delta\boldsymbol{w}(\boldsymbol{s},\boldsymbol{s}';\sigma_d,\sigma_t) = G(\Delta_{d,i}^{(s,s')},\sigma_d) \cdot G(\Delta_{t,i}^{(s,s')},\sigma_t)$. A modified {\small \texttt{analyzeTrajectories}} procedure for Algorithm \ref{algorithm_reachability_summary} implementing the equation above is presented in Algorithm \ref{algorithm_modified_createEPairs} in Appendix Section \ref{appendix:createAPairs}.

\subsection{Scalability of Algorithm \ref{algorithm_reachability_summary}}\label{scalability_of_algorithm}
\begin{figure}
    \centering
    \begin{subfigure}{0.24\textwidth}
        \centering
        \includegraphics[width=\textwidth]{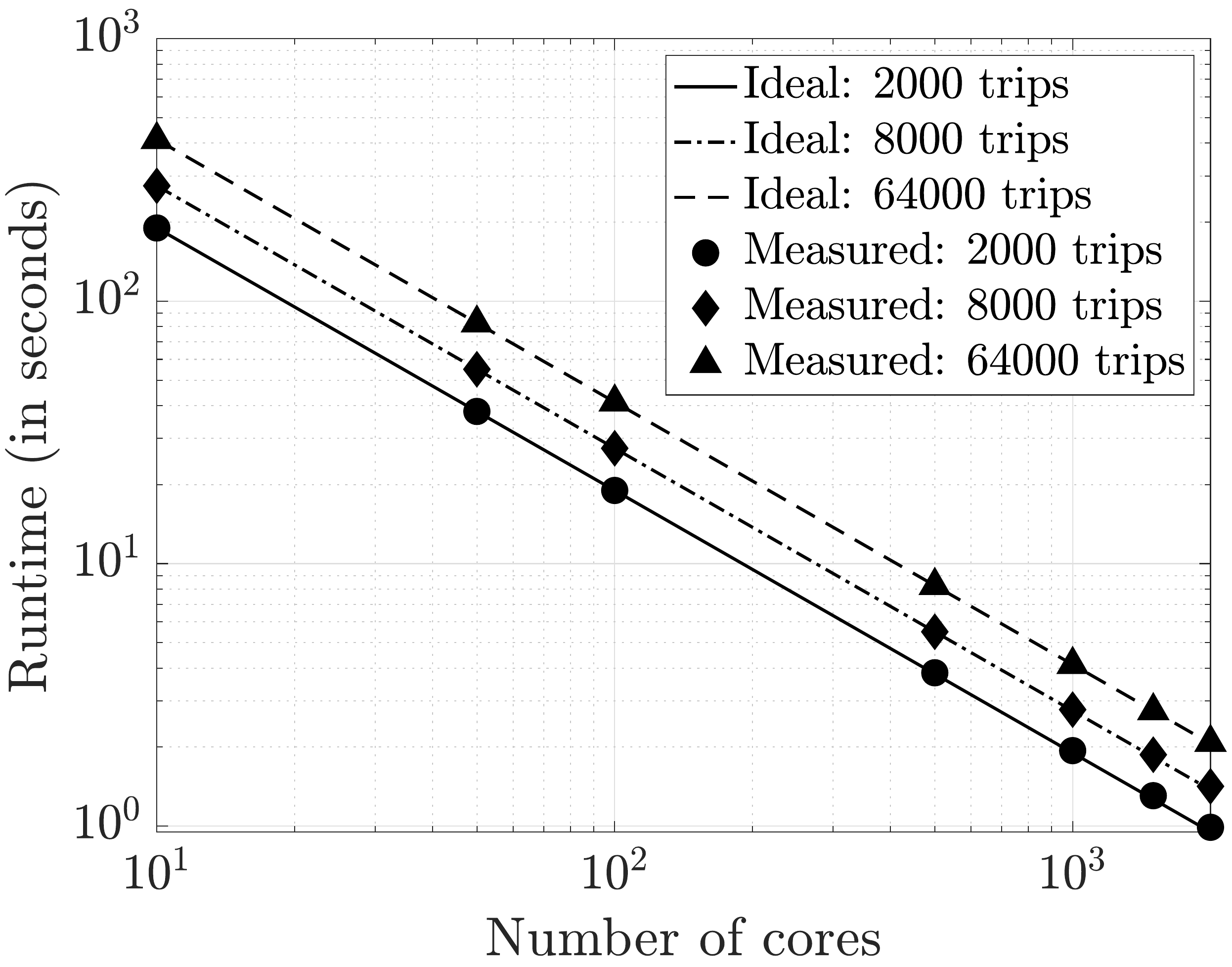}
        \caption{}
        \label{scaling_analysis_runtime}
    \end{subfigure}
    \hfill
    \begin{subfigure}{0.24\textwidth}
        \centering
        \includegraphics[width=\textwidth]{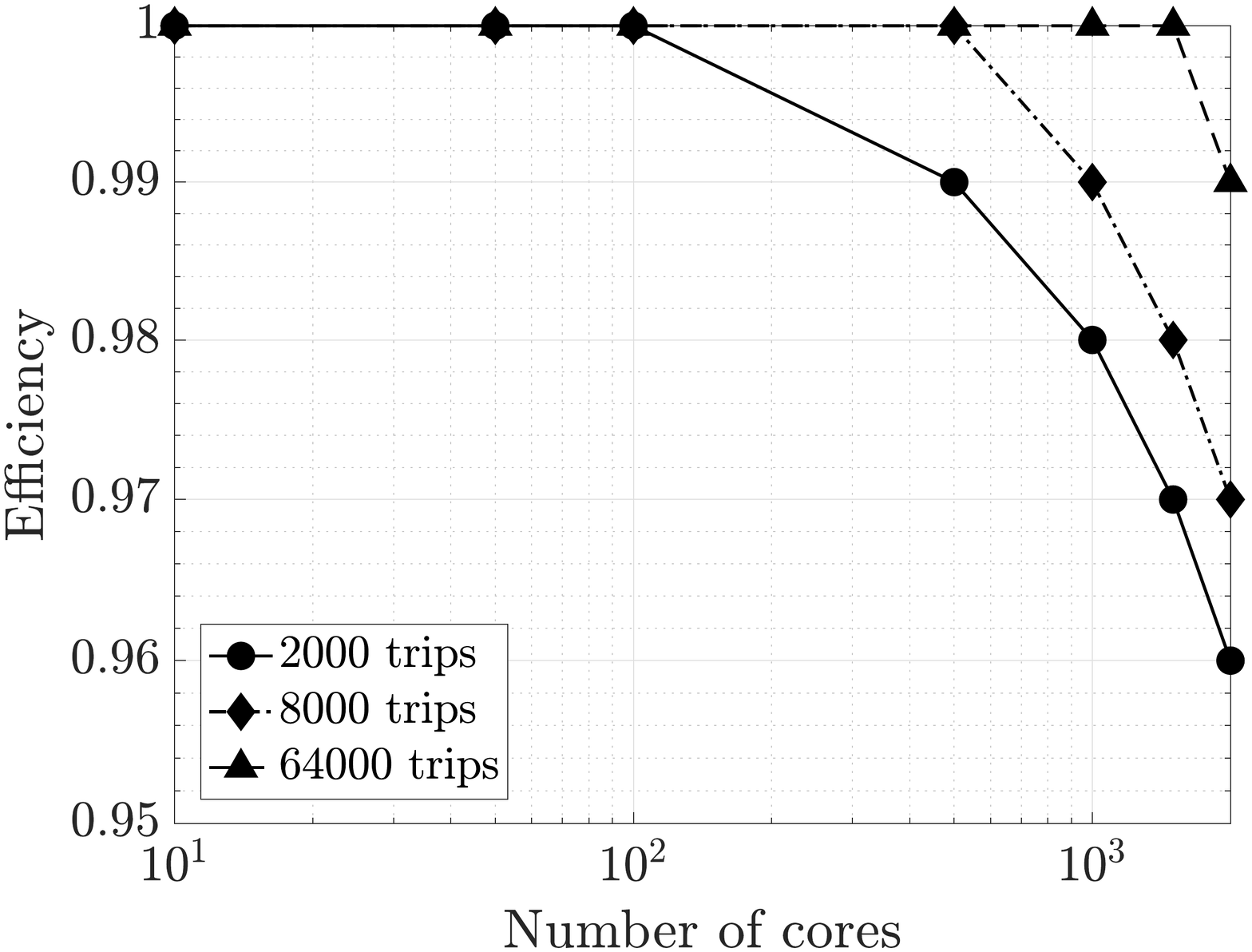}
        \caption{}
        \label{scaling_analysis_efficiency}
    \end{subfigure}
    \hfill
    \begin{subfigure}{0.24\textwidth}
        \centering
        \includegraphics[width=\textwidth]{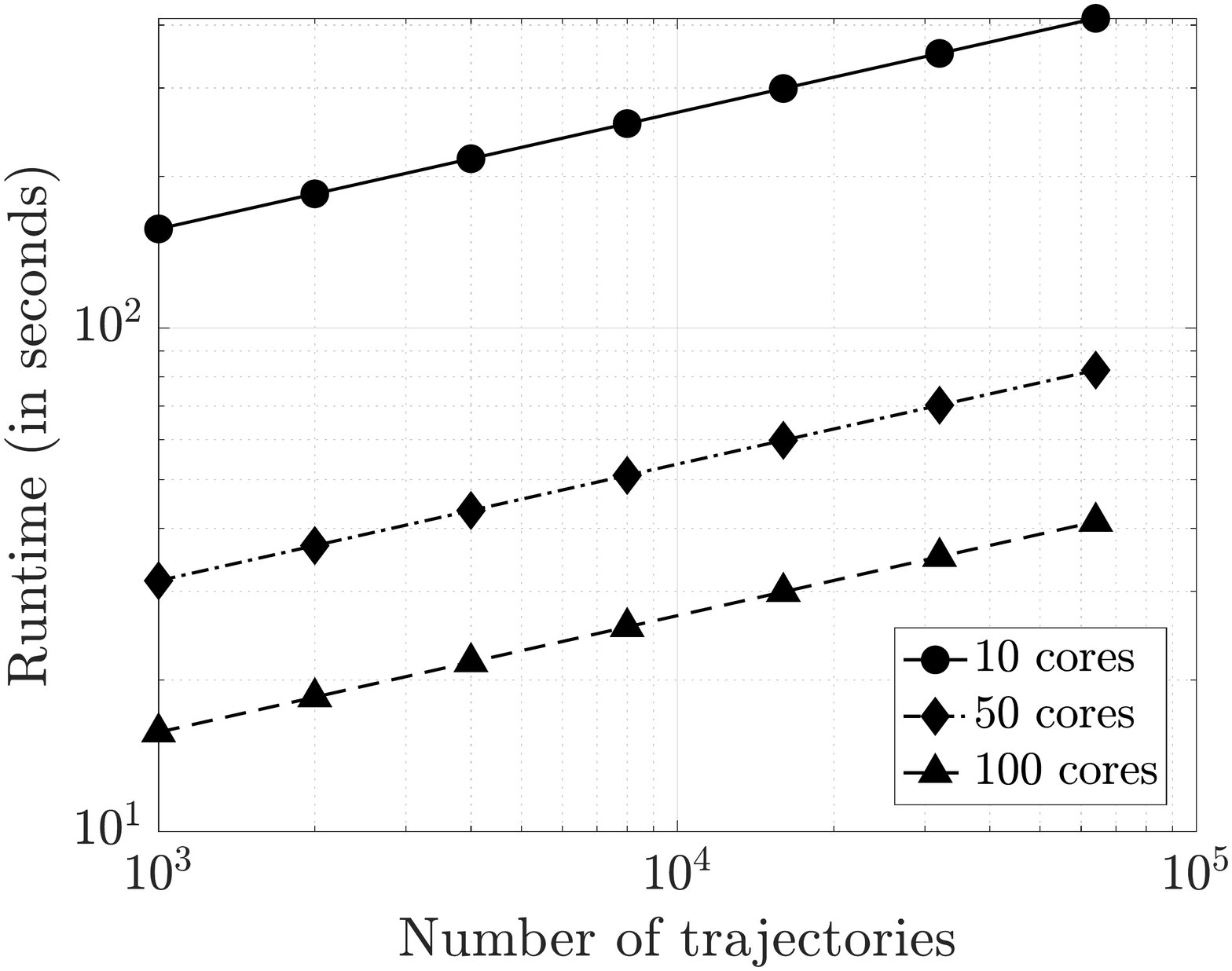}
        \caption{}
        \label{scaling_analysis_trajectories}
    \end{subfigure}
    \caption{Strong scaling analysis of (a) runtime, (b) efficiency, and (c) scaling with number of trajectories of Algorithm \ref{algorithm_reachability_summary}.}
    \label{scaling_analysis}
\end{figure}

Algorithm \ref{algorithm_reachability_summary} modified using Algorithm \ref{algorithm_modified_createEPairs} is implemented in a distributed data-parallel fashion using the Scala-based, Spark Dataset API. To demonstrate scalability, the publicly available T-Drive dataset \cite{msrtaxi11} is suitably pre-processed (detailed in Appendix Section \ref{tdrive_preprocessing}) to obtain three datasets of 2000, 8000, and 64000 trips on which strong scaling \cite{grama03} analysis is performed. The Spark implementation is executed on these datasets by varying the number of CPU cores from 10 to 2000. The mean value of the runtime and strong scaling efficiency\footnote{For a given dataset size, if $t_{c}^{\text{id}}$ is the ideal time required on $c$ CPU cores, $t_{c_0}^{\text{m}}$ is the time measured when executing algorithm on $c_0$ cores, and $e_{ss}$ is strong scaling efficiency, then $t_{c}^{\text{id}} = t_{c_0}^{\text{m}}(c_0/c)$ and $e_{ss} = t_{c}^{\text{id}}/t_{c_0}^{\text{m}}$.} \cite{grama03} obtained over seven runs is plotted in Figures \ref{scaling_analysis_runtime} and \ref{scaling_analysis_efficiency}, respectively, as the number of CPU cores are varied for the three datasets. Typical characteristics of parallel applications such as increased efficiency on larger number of cores for bigger problem sizes, efficiency drop for smaller problem sizes as number of cores increases, and communication latency dominating compute time are observed.

\setlength{\textfloatsep}{0pt}
\begin{algorithm}
    \small
    \caption{Generate $\boldsymbol{\Psi}_{s}^{ea}$ for all nodes $\boldsymbol{s} \in \overline{\boldsymbol{V}}_{ES}$}\label{algorithm_reachability_summary}
    \begin{algorithmic}
        \INPUT $\mathcal{T}_{t_0}^{t_0 + \Delta t} = (\boldsymbol{T}_1 \ldots \boldsymbol{T}_M)$, $|\boldsymbol{T}_i| = n_i \,\,\forall\,\, i \in [1,M]$, $\boldsymbol{T}_i = (\boldsymbol{p}_1^i, \ldots, \boldsymbol{p}_{n_i}^i)$ where $\boldsymbol{p}_k^i = (\boldsymbol{z}_k^i, \boldsymbol{t}_k^i)$
        \OUTPUT Map $\boldsymbol{\Psi}: \boldsymbol{s} \rightarrow \boldsymbol{\Psi}_{s}^{ea}$
        \PARAMETERS $\boldsymbol{\delta}_R$. Define $\boldsymbol{L} = 2\boldsymbol{\delta_R}+1$. Let $\overline{n} = \max\{n_1, \ldots, n_M\}$
        \Procedure{ReachabilitySummaryGenerator}{$\phantom{}$}
            \State Initialize map $\boldsymbol{\Psi} = \emptyset$
            \State $\boldsymbol{S} \leftarrow \texttt{analyzeTrajectories}(\mathcal{T}_{t_0}^{t_0 + \Delta t})$
            \For{node $\boldsymbol{s} \in \boldsymbol{S}$}
                \State Arrays $\boldsymbol{\Psi}_{s}^{a}$, $\boldsymbol{\Psi}_{s}^{e}$. Zero-initialized of length $\boldsymbol{L}^2$  
                \State $\boldsymbol{S}_{s}^{e},\,\boldsymbol{S}_{s}^{a} \leftarrow \boldsymbol{S}[\boldsymbol{s}]$ 
                \State $\boldsymbol{\Psi}_{s}^{e}[\kappa] = \boldsymbol{S}_{s}^{e}[\kappa]$ \textbf{for all} keys $\kappa$ of map
                $\boldsymbol{S}_{s}^{e}$
                \State $\boldsymbol{\Psi}_{s}^{a}[\kappa] = \boldsymbol{S}_{s}^{a}[\kappa]$ \textbf{for all} keys $\kappa$ of map $\boldsymbol{S}_{s}^{a}$
                \State Reshape $\boldsymbol{\Psi}_{s}^{e}$ and $\boldsymbol{\Psi}_{s}^{a}$ to $\boldsymbol{L}\times\boldsymbol{L}$ arrays
                \State $\boldsymbol{\Psi}_{s}^{ea} \leftarrow$ Stack channels $\boldsymbol{\Psi}_{s}^{e}$, $\boldsymbol{\Psi}_{s}^{a}$
                \State $\boldsymbol{\Psi}[\boldsymbol{s}] \leftarrow \boldsymbol{\Psi}_{s}^{ea}$
            \EndFor
        \EndProcedure
        \Function{analyzeTrajectories}{$\mathcal{T}_{t_0}^{t_0 + \Delta t}$}\Comment{$\mathcal{O}(M\overline{n}^2)$}
            \State Initialize map $\boldsymbol{S} = \emptyset$\
            \For{trajectory $\boldsymbol{T}_i \in \mathcal{T}_{t_0}^{t_0 + \Delta t}$}
                \For{$k \leftarrow 1$ to $n_i$}
                    \For{$l \leftarrow k$ to $n_i$}
                        \If{$\boldsymbol{z}_l^i \in \boldsymbol{N}_{z_k^i}^{{\delta_R}}$}
                            \State if $\boldsymbol{z}_k^i \notin \boldsymbol{S}$, $\boldsymbol{S}[\boldsymbol{z}_k^i] = (\boldsymbol{S}_{z_k^i}^{e} = \emptyset, \boldsymbol{S}_{z_k^i}^{a} = \emptyset)$
                            \State if $\boldsymbol{z}_l^i \notin \boldsymbol{S}$, $\boldsymbol{S}[\boldsymbol{z}_l^i] = (\boldsymbol{S}_{z_l^i}^{e} = \emptyset, \boldsymbol{S}_{z_l^i}^{a} = \emptyset)$
                            \State $r_{z_k^i}(z_l^i) \leftarrow \texttt{getIndex}(\boldsymbol{z}_k^i, \boldsymbol{z}_l^i)$
                            \State $r_{z_l^i}(z_k^i) \leftarrow \texttt{getIndex}(\boldsymbol{z}_l^i, \boldsymbol{z}_k^i)$
                            \State $c_{z_k^i}^{a}(z_l^i) = c_{z_l^i}^{e}(z_k^i) = 1.0$\Comment{Refined in III(C)}
                            \State $\boldsymbol{S}_{z_k^i}^{a}[r_{z_k^i}(z_l^i)] \pluseq c_{z_k^i}^{a}(z_l^i)$
                            \State $\boldsymbol{S}_{z_l^i}^{e}[r_{z_l^i}(z_k^i)] \pluseq c_{z_l^i}^{e}(z_k^i)$
                        \EndIf
                    \EndFor
                \EndFor
            \EndFor
            \State \Return $\boldsymbol{S}$
        \EndFunction
        \Function{getIndex}{$\boldsymbol{s}$,$\boldsymbol{s}'$}\Comment{$\mathcal{O}(1)$}
            \State \Return $\boldsymbol{L}(\boldsymbol{y}_{s'}-\boldsymbol{y}_{s} + \boldsymbol{\delta}_R) + (\boldsymbol{x}_{s'}-\boldsymbol{x}_{s} + \boldsymbol{\delta}_R)$
        \EndFunction
    \end{algorithmic}
\end{algorithm}

Another scalability test is to increase the number of trajectories processed on a fixed number of cores. Plotting the runtime of Algorithm \ref{algorithm_reachability_summary} as the number of trajectories are varied from 1000 to 64000 and the number of cores are varied from 10 to 100 in Figure \ref{scaling_analysis_trajectories}, it is empirically observed that the algorithm run-time scales sub-linearly with the number of trajectories with a power law exponent of roughly 1/5. While Figure \ref{scaling_analysis} and the observations in this sub-section are stated specifically for the T-Drive dataset, similar observations are empirically observed on much larger proprietary datasets (described in section \ref{dataset_description}). Scalability of Algorithm \ref{algorithm_reachability_summary} is especially crucial in a setting where embeddings need to be generated periodically for maintaining and updating geospatial maps given the large number of zoom-24 tiles that are possible on the surface of the earth. Contrast this to the vocabulary sizes ($\mathcal{O}(10^{6})$) used in NLP approaches like word2vec \cite{mikolov13}. From a machine learning systems perspective \cite{trinity21}, scalable generation of reachability embeddings makes it convenient for feature stores \cite{featurestore} to store generated embeddings as reusable features for training and deployment of multiple downstream task-specific models.

\subsection{Encoding reachability summaries}\label{encoding_reachability_summaries}
The neural architecture of the custom-designed, fully-convolutional, contractive autoencoder, $\mathcal{F}_{R}$, is shown in Figure \ref{contractive_autoencoder} of Appendix Section \ref{contractive_autoencoder_architecture}. During training, the encoder, $\mathcal{F}_{R}^{e}$, maps $\boldsymbol{\Psi}_{s}^{ea}$ to a compressed representation, $\boldsymbol{h}(\boldsymbol{s}) = \mathcal{F}_{R}^{e}(\boldsymbol{\Psi}_{s}^{ea}) \in \mathbb{R}^{\boldsymbol{d}_R}$, which is then used by the decoder, $\mathcal{F}_{R}^{d}$, to reconstruct $\boldsymbol{\Psi}_{s}^{ea}$. The objective function to be minimized, $\mathcal{L}$, consists of the reconstruction loss, $\mathcal{L}_{rec}$, regularized by the contractive \cite{rifai11} loss, $\mathcal{L}_{con}$, shown in equations (\ref{reconstruction_loss}), (\ref{contractive_loss}), and (\ref{total_loss}).
\begin{subequations}\label{loss_functions}
    \begin{gather}
        \mathcal{L}_{rec}(\boldsymbol{s};\boldsymbol{\theta}) = \norm{\boldsymbol{\Psi}_{s}^{ea} - \mathcal{F}_{R}^{d}(\mathcal{F}_{R}^{e}(\boldsymbol{\Psi}_{s}^{ea}))}_{2}^{2} \label{reconstruction_loss}\\
        \mathcal{L}_{con}(\boldsymbol{s};\boldsymbol{\theta}) = \sum_{i=0}^{d_R-1} \norm{\nabla_{\Psi_s^{ea}} \boldsymbol{h}_{i}(\boldsymbol{s})}_{2}^{2} \label{contractive_loss}\\
        \mathcal{L}(\boldsymbol{\theta}) = \mathbb{E}_{\boldsymbol{s}\sim\overline{\boldsymbol{V}}_{ES}}\left[\mathcal{L}_{rec}(\boldsymbol{s};\boldsymbol{\theta}) + \lambda\mathcal{L}_{con}(\boldsymbol{s};\boldsymbol{\theta})\right] \label{total_loss}
    \end{gather}
\end{subequations}
Here, $\boldsymbol{\theta}$ are the parameters of $\mathcal{F}_{R}$ and $\lambda$ is a hyperparameter. A ReLU nonlinearity forces all embeddings to be positive real vectors. Since there is no upper bound on the pixel values in $\boldsymbol{\Psi}_{s}^{ea}$, the input to $\mathcal{F}_{R}$ is log-normalized and the reconstruction output exponentiated (similar to \cite{xiao20}) before calculating the losses (\ref{reconstruction_loss})--(\ref{total_loss}) to train the neural network stably and avoid spurious effects from the large dynamic range of $\boldsymbol{\Psi}_{s}^{ea}$. The training set is composed of zoom-24 tiles from a variety of geographically diverse regions. Once trained, $\mathcal{F}_{R}^{e}$ is used to generate the embeddings, $\boldsymbol{h}(\boldsymbol{s})$, for all $\boldsymbol{s} \in \overline{\boldsymbol{V}}_{ES}$. As shown in \cite{rifai11}, contractive regularization forces $\boldsymbol{h}(\boldsymbol{s})$ to be invariant to small perturbations of $\boldsymbol{\Psi}_{s}^{ea}$ (locally contractive in neighborhood of $\boldsymbol{\Psi}_{s}^{ea}$) thereby yielding robust and sparse compressed representations of $\boldsymbol{\Psi}_{s}^{ea}$. Presence of $\mathcal{L}_{con}$ yields only a few non-zero elements in the embeddings for most nodes. From the perspective of energy-based models, $\mathcal{L}_{con}$ helps minimize the volume of low-energy regions to help learn semantically meaningful representations.

\setlength{\belowcaptionskip}{-2pt}
\setlength{\textfloatsep}{5pt}

\section{Dataset Description}\label{dataset_description}
Small, publicly available GPS trajectory datasets (e.g., \cite{msrtaxi11},\cite{ucigpsdataset13},\cite{osmgpstraces21}, \cite{yuzheng2009}) (i) have varying sampling rates with incomplete trajectories, (ii) do not cover the entire road network in the geographical areas they are obtained from, (iii) have inconsistent observation times and intervals, and (iv) lack geographical diversity. Most importantly, in order to test the utility of learned unsupervised representations like reachability embeddings on downstream supervised tasks, we need labeled data for these tasks. To the best of our knowledge, there is no sufficiently large, geographically diverse, publicly available GPS trajectory dataset covering a sufficiently large observation time interval which can be matched to labeled datasets collected for various geospatial tasks in the same geographical regions.

To overcome this difficulty, we use \textit{probe data} \cite{tcrunch18}, which is a privacy-preserving, structured, sequential, and proprietary dataset\footnote{Probe data is similar to publicly available GPS trajectory datasets such as \cite{ucigpsdataset13,msrtaxi11}. Section titled ``Probe data and privacy'' in \cite{tcrunch18} has more details.}. Labels are collected across a variety of geographic regions, in the form of \textit{Well-Known Text} (WKT) polygons\footnote{Geospatial feature coordinates are suitably buffered to obtain WKT polygons. Assigning value 1 to pixels inside and 0 to those outside the polygons yields labels for semantic segmentation.} \cite{wikiwkts21}, for training supervised semantic segmentation models for five downstream geospatial tasks described in Section \ref{experiments_and_results}.

There are multiple ways, called \textit{local aggregate representations} (LAR), simpler to implement than reachability embeddings, to create semantically meaningful, multi-channel, \textit{image-like} representations of GPS trajectory datasets making them amenable for use by downstream computer vision models for geospatial tasks. The pixel values for a given zoom-24 tile are obtained by analyzing records observed \textit{only in that tile} (local) to yield one (single-channel) or a vector of aggregate counts (multi-channel). Three such representations are count-based raster map (CRM), heading count-based raster map (HCRM), and speed count (SC).

Given the Cartesian grid structure of $\boldsymbol{V}_{ES}$, associating $\boldsymbol{c}$ meaningful real numbers (represented as $\boldsymbol{c}$ channels) to each tile (pixel) within any set of contiguous $\boldsymbol{h} \times \boldsymbol{w}$ tiles (pixels) forms a $\boldsymbol{h} \times \boldsymbol{w} \times \boldsymbol{c}$-dimensional image-like representation of the geographic location represented by the $\boldsymbol{h} \times \boldsymbol{w}$ tiles, as shown in Figure \ref{image_like_representation}. A \textit{count-based raster map} (CRM) is a single-channel representation ($\boldsymbol{c} = 1$) where the value of each pixel is the number of occurrences of GPS records in the zoom-24 tile corresponding to the pixel counted over all trajectories in $\boldsymbol{\mathcal{T}}(t_0, \Delta t)$. Instead, if the occurrences of records per pixel is bucketed based on the direction in which the record was heading into 12 buckets of $30^{\circ}$, and each bucket is represented as an individual channel ($\boldsymbol{c} = 12$), we obtain the \textit{heading count-based raster map} (HCRM) representation. Similarly, if the occurrences of records per pixel are bucketed based on the speed of the GPS records into 14 buckets of 5 miles per hour starting from 0, and each bucket is represented as an individual channel ($\boldsymbol{c} = 14$), we obtain the \textit{speed-count} (SC) representation. 

If pixel values are set to the reachability embeddings of the corresponding zoom-24 tile, we obtain the $\boldsymbol{d}_R$-channel reachability embeddings representation. Reachability embeddings may be viewed as a mapping of a geographic location to an \textit{embedding field} with $\boldsymbol{h} \times \boldsymbol{w} \times \boldsymbol{d}_R$ parameters based on observed mobility in that geographic location over a specified observation interval $\Delta t$. In contrast to LAR, reachability embeddings are \textit{global} pixel-level representations --- representation of tile $\boldsymbol{s}$ depends on all activity in $\boldsymbol{N}_{s}^{\delta_R}$. Trajectory-independent, road geometry related information is supplied using the \textit{road network presence} (RNP) channel ($\boldsymbol{c}=1$) which is a binary representation of the road network as a raster image where each pixel is assigned the value 1 if the corresponding zoom-24 tile has a road segment present and the value 0 otherwise. In this paper, $\boldsymbol{h} = \boldsymbol{w} = 256$.

Analogous to spectral imaging for remote sensing, where channels corresponding to different spectral bands capture additional information of a geographic location to complement their RGB (visible spectrum) counterparts, CRM, HCRM, SC, and reachability-based channels are multi-channel, image-like representations of information of the location deduced from trajectory data. The RNP channel can be considered as an \textit{ontological} representation (where traffic \textit{should} exist) of $\boldsymbol{G}_{ES}$ while trajectory datasets are the \textit{epistemological} representation (where traffic \textit{actually} exists) of $\boldsymbol{G}_{ES}$.

\section{Experiments and Results}\label{experiments_and_results}
The impact of reachability embeddings is evaluated on the performance of supervised, pixel-wise, semantic segmentation models for five downstream geospatial tasks, viz. (i) detection of overpasses, (ii) detection of pedestrian crosswalks, (iii) detection of driving access (entry/exit) points, (iv) detection of locations with traffic lights, and (v) detection of locations with stop signs. For all tasks, the UNet \cite{ronneberger2015} architecture is trained to minimize the pixel-wise binary cross-entropy between the predicted segmentation map and labels. A 60-20-20 randomly chosen training-validation-testing set is created and fixed for all subsequent experimentation.

Since reachability embeddings encode spatial connectivity (frequency and directionality of transitions) and speed (distance covered and time taken), baseline models using a combination (best model obtained after ablation study) of LAR-based channels as inputs are compared to the same model trained by replacing all LAR-based channels with reachability embeddings-based channels. Direct comparison between (i) models using inputs accounting for spatial connectivity versus inputs that do not, and (ii) the proposed self-supervised, learned representations versus explicit encoding of different aspects of trajectory data, demonstrates that learning more informative representations like reachability embeddings lead to better performance on downstream supervised tasks while using lesser trajectory data (i.e., reduced observation intervals, $\Delta t$). Probe data-based trajectory sets filtered for driving ($\boldsymbol{\mathcal{T}}^{d}(t_0,\Delta t)$) and walking ($\boldsymbol{\mathcal{T}}^{w}(t_0,\Delta t)$) motion modalities are obtained. Reachability embeddings (RE) are generated for both walking and driving modalities and denoted by WRE and DRE, respectively. CRM is also computed for both walking and driving modalities and denoted by WCRM and DCRM, respectively. HCRM and SC are calculated only for driving modality. For the models built to evaluate the predictive power of reachability embeddings on downstream tasks (results in table \ref{results_table}), only trajectory-based representations and/or RNP are used as inputs. No inputs from other remote sensing modalities (e.g., satellite or SAR imagery) are used since the goal of these experiments is to assess image-like representations of trajectory data for solving geospatial tasks as opposed to building the best model (which may combine multimodal inputs) for a given task. The last part of this section (results in table \ref{multimodal_results_table}) deals with multimodal modeling with reachability embeddings.

\setlength{\belowcaptionskip}{-2pt}
\setlength{\textfloatsep}{5pt}
\begin{table}[hbt!]
    \scriptsize
    \caption{Combination of LAR-based channels identified from ablation study for the best baseline model for the downstream tasks. Corresponding inputs for reachability embeddings (RE) based models. \label{baselines_table}}
    \centering
    \begin{tabular}{X Y Z Z Z Y Y Y Y}
        \toprule
                        & \multicolumn{5}{c}{Baseline Models}                                      & \multicolumn{3}{c}{Models w/ RE}           \\
        \cmidrule(lr){2-6}\cmidrule(lr){7-9}
        Application     & RNP          & DCRM         & WCRM         & HCRM         & SC           & RNP          & DRE          & WRE          \\
        \midrule
        Overpass        & $\times$     & $\times$     & $\times$     & $\checkmark$ & $\times$     & $\times$     & $\checkmark$ & $\times$     \\
        Crosswalk       & $\times$     & $\checkmark$ & $\checkmark$ & $\times$     & $\times$     & $\times$     & $\checkmark$ & $\checkmark$ \\
        Access Point    & $\checkmark$ & $\times$     & $\times$     & $\checkmark$ & $\times$     & $\checkmark$ & $\checkmark$ & $\times$     \\
        Traffic Lights  & $\checkmark$ & $\checkmark$ & $\times$     & $\checkmark$ & $\checkmark$ & $\checkmark$ & $\checkmark$ & $\times$     \\
        Stop Signs      & $\checkmark$ & $\checkmark$ & $\checkmark$ & $\checkmark$ & $\checkmark$ & $\checkmark$ & $\checkmark$ & $\checkmark$ \\
        \bottomrule
    \end{tabular}
\end{table}

\subsection{Baselines for Each Downstream Task} 
An ablation study is performed over all 31 possible combinations of RNP, DCRM, WCRM, HCRM, and SC to identify the best combination of LAR-based input channels to build the baseline model for each task. Each input combination is concatenated along the channel dimension. Inputs are log-normalized so as to avoid spurious effects of the large dynamic range owing to the lack of an upper bound on the pixel-values. Hyperparameters are tuned on the validation set. Once trained, segmentation map predictions from all 31 models for each task is obtained for all examples of the test set. For every task, using ground truth labels available for the test set, the mean precision-recall curve (PRC) and the area under the curve (AUPRC) over all test examples is obtained for each of the 31 models by varying the confidence threshold of the binary segmentation outputs. The combination of LAR-based input representations that yield the highest AUPRC for each task is chosen as the baseline model for that task and is documented in Table \ref{baselines_table}.

\setlength{\belowcaptionskip}{0pt}
\begin{table*}[ht]
    \footnotesize
    \caption{Comparison of AUPRC ($\uparrow$ is better) obtained from both variants of LAR and reachability-based models for 5 downstream geospatial tasks. Percentage gain compared to first row shown in brackets. \label{results_table}}
    \centering
    \begin{tabular}{U T V V V V V}
        \toprule
        Input Channels                      & Observation Interval & Overpass Detection       & Crosswalk Detection     & Access Point Detection   & Traffic Lights Detection & Stop Signs Detection     \\
        \midrule
        LAR Baseline                        & $\Delta t$           & 0.782                    & 0.922                   & 0.663                    & 0.890                    &  0.921                   \\  
        LAR Baseline                        & $3\Delta t$          & 0.785 (+0.4\%)           & 0.923 (+0.1\%)          & 0.684 (+3.0\%)           & 0.925 (+3.8\%)           &  0.924 (+0.3\%)          \\ 
        Reachability, $\boldsymbol{d}_R=8$  & $\Delta t$           & 0.899 (+15.0\%)          & 0.959 (+4.0\%)          & 0.843 (+21.4\%)          & 0.974 (+8.6\%)           &  0.938 (+1.9\%)          \\ 
        Reachability, $\boldsymbol{d}_R=16$ & $\Delta t$           & \textbf{0.925} (+18.3\%) & \textbf{0.976} (+5.9\%) & \textbf{0.864} (+23.3\%) & \textbf{0.985} (+9.6\%)  &  \textbf{0.959} (+4.1\%) \\
        \bottomrule
    \end{tabular}
\end{table*}

\setlength{\belowcaptionskip}{-2pt}
\setlength{\textfloatsep}{5pt}
\begin{figure*}
    \centering
    \begin{subfigure}{0.19\textwidth}
        \centering
        \includegraphics[width=\textwidth]{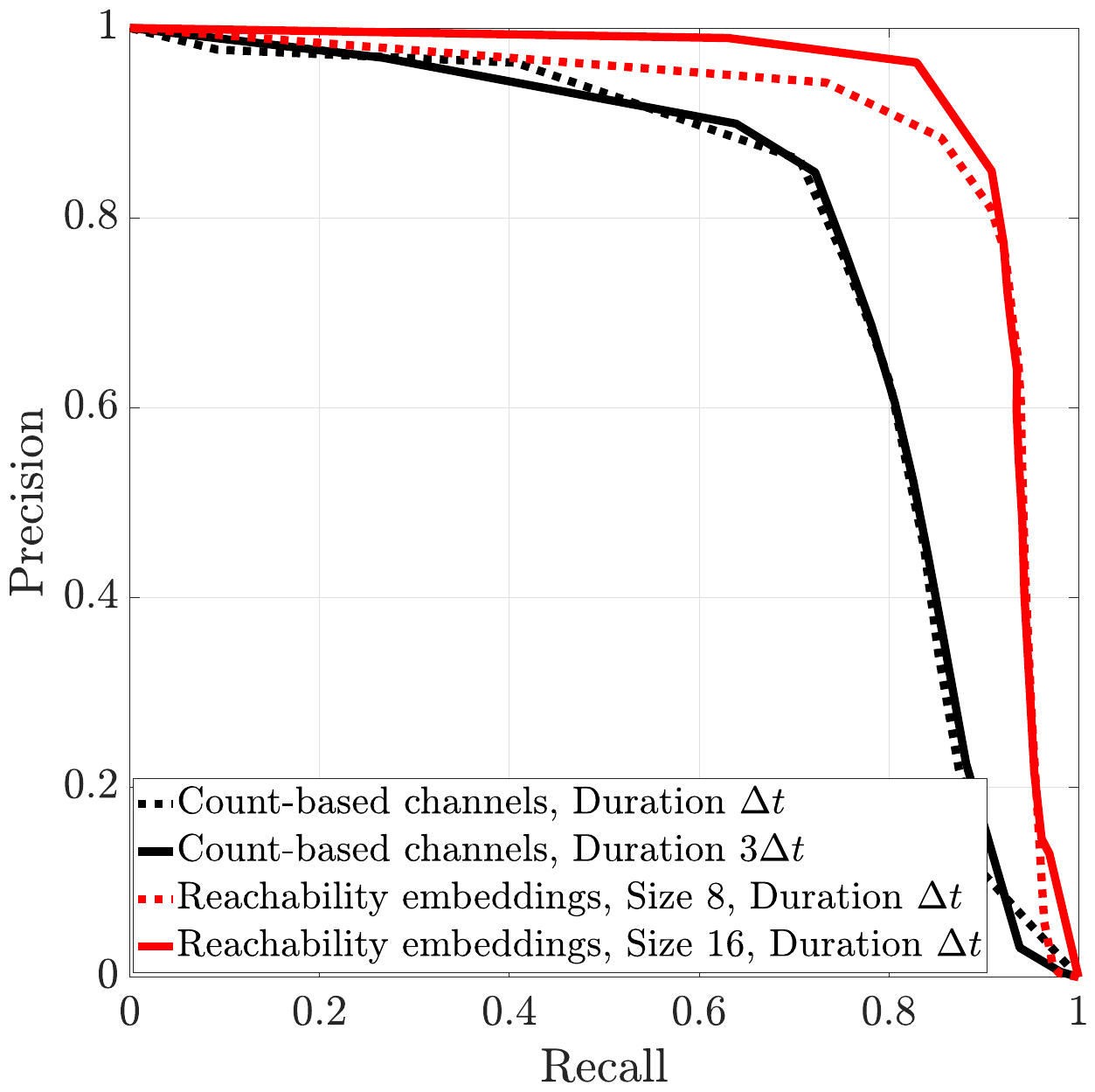}
        \caption{Overpass}
        \label{prcurves:multiz}
    \end{subfigure}
    \hfill
    \begin{subfigure}{0.19\textwidth}
        \centering
        \includegraphics[width=\textwidth]{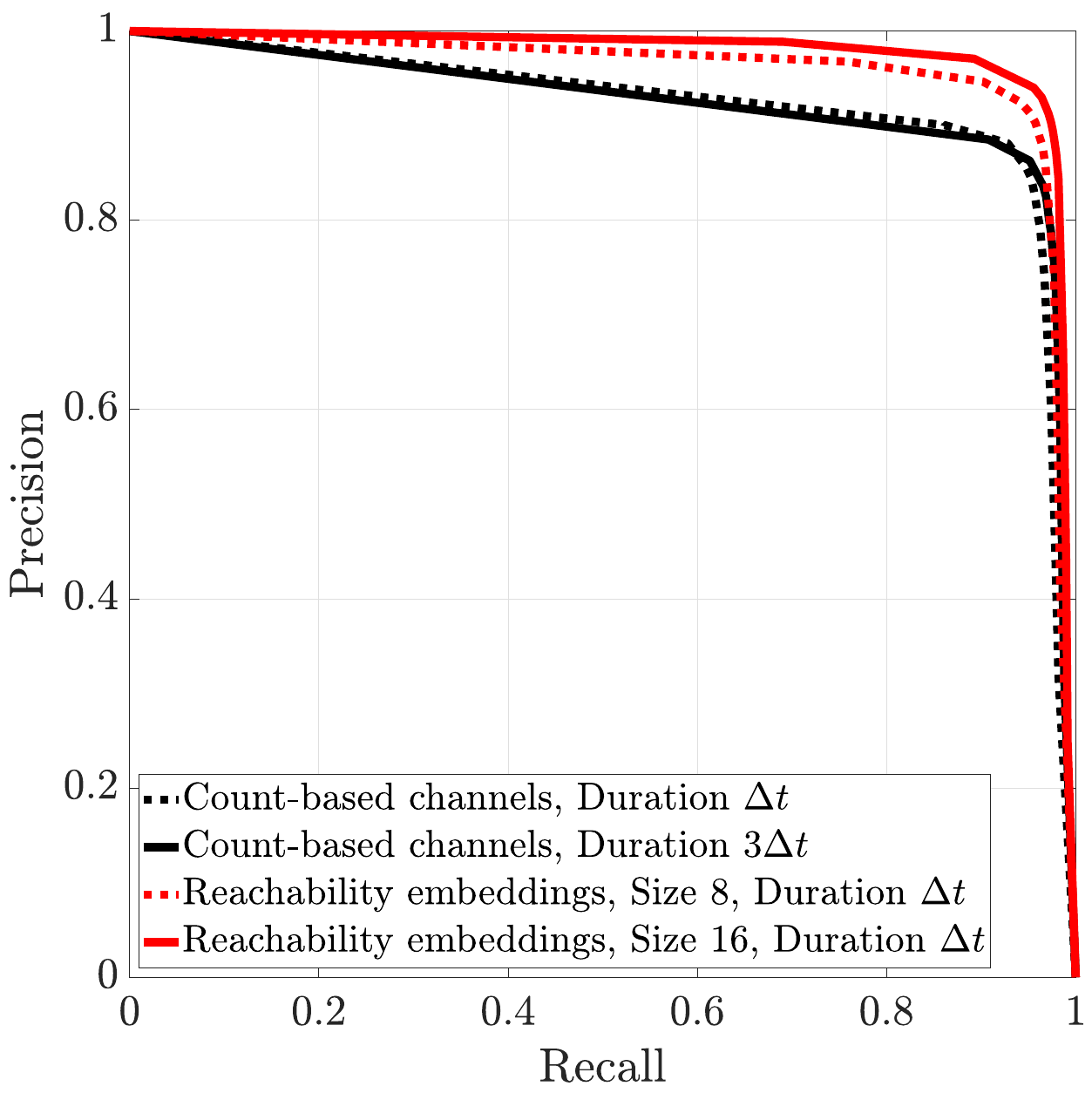}
        \caption{Crosswalk}
        \label{prcurves:pedxing}
    \end{subfigure}
    \hfill
    \begin{subfigure}{0.19\textwidth}
        \centering
        \includegraphics[width=\textwidth]{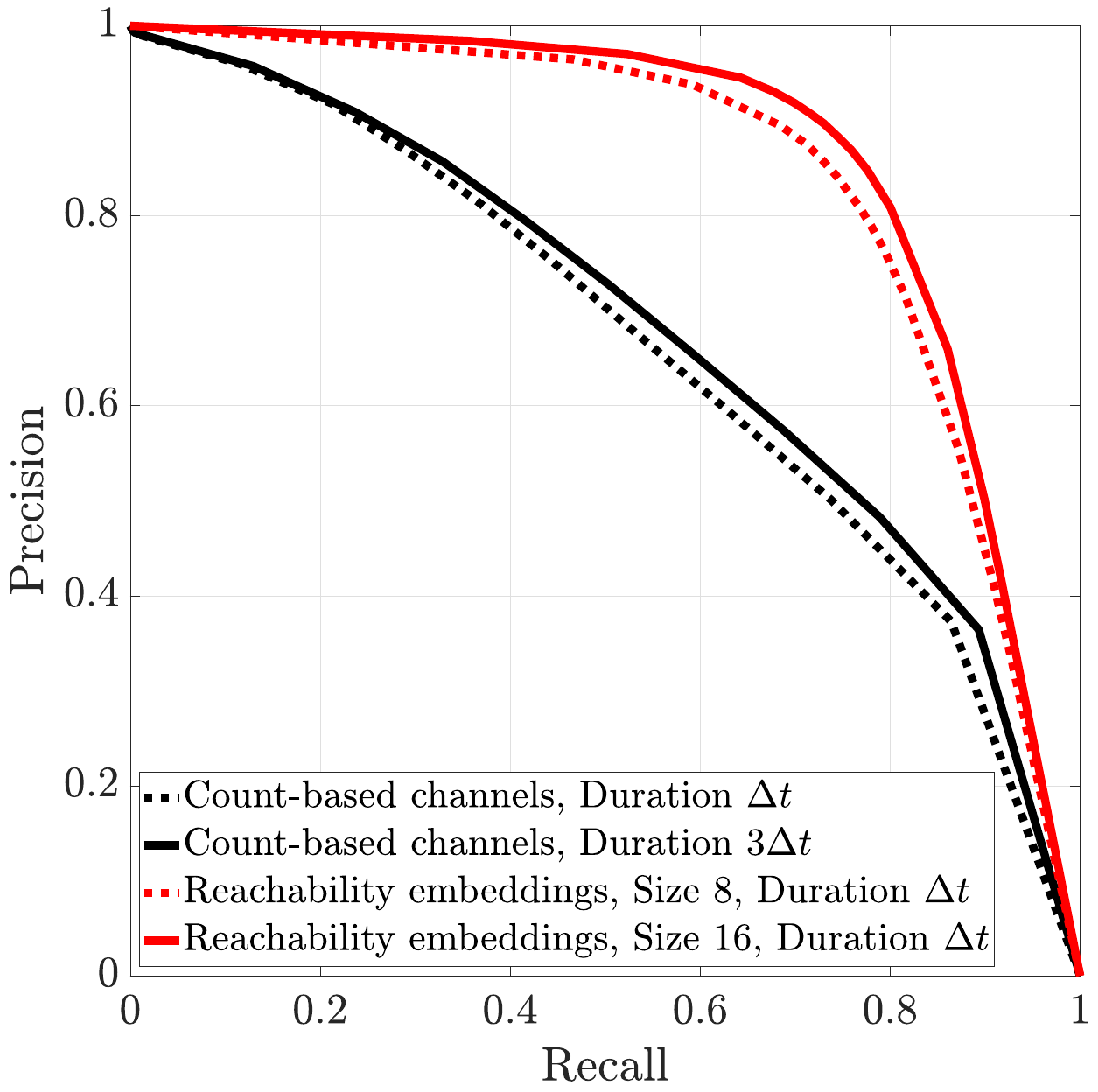}
        \caption{Access Points}
        \label{prcurves:draps}
    \end{subfigure}
    \hfill
    \begin{subfigure}{0.19\textwidth}
        \centering
        \includegraphics[width=\textwidth]{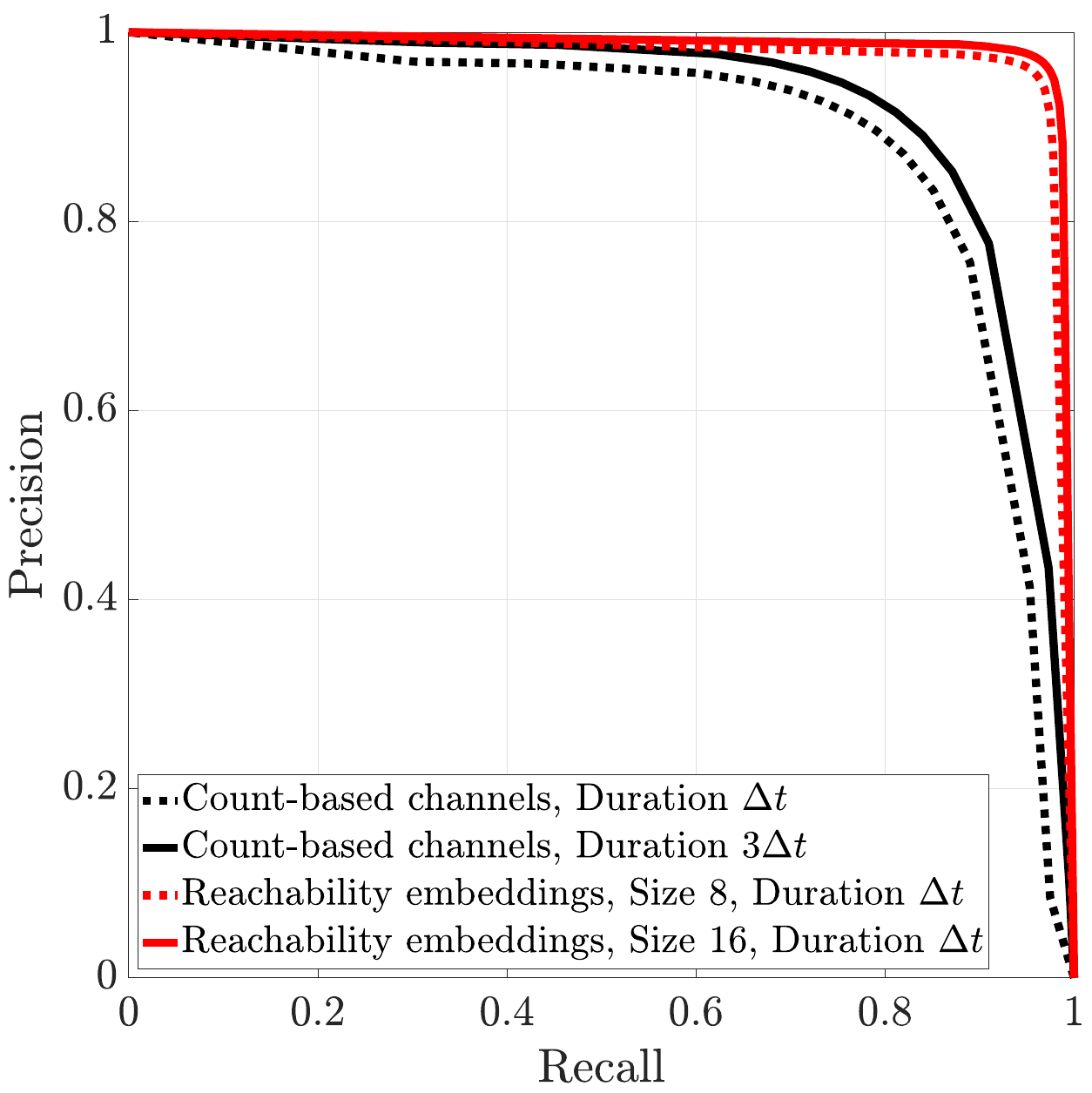}
        \caption{Traffic Lights}
        \label{prcurves:trafficlights}
    \end{subfigure}
    \hfill
    \begin{subfigure}{0.19\textwidth}
        \centering
        \includegraphics[width=\textwidth]{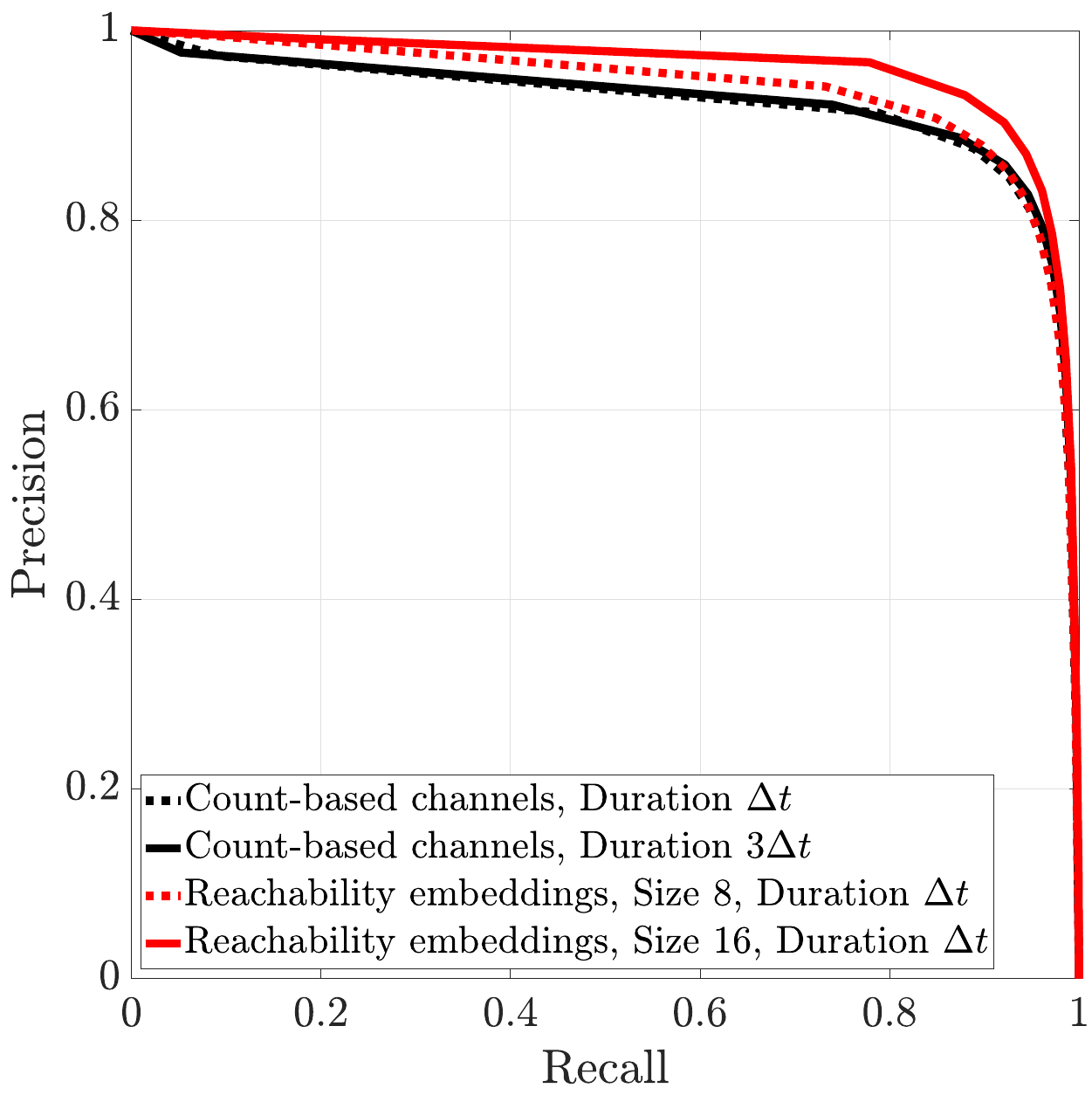}
        \caption{Stop Signs}
        \label{prcurves:stopsigns}
    \end{subfigure}
    \caption{Precision-Recall curves obtained from both variants of LAR and reachability-based models for the 5 downstream tasks.}
    \label{prcurves}
\end{figure*}

\subsection{Experiments with Reachability Embeddings} 
With the goal of assessing the importance of self-supervised representations that account for various aspects of spatial connectivity (transition frequency, time, and distance), reachability-based models, documented in Table \ref{baselines_table}, are built that directly correspond to the baseline models by simply replacing any combination of DCRM, HCRM, and SC channels with DRE and replacing WCRM with WRE. Two variants each of the reachability-based models and baseline models are compared: reachability-based models that use DRE and WRE computed from trajectories observed within $[t_0, t_0 + \Delta t]$ with varying embedding sizes (a) $\boldsymbol{d}_R=8$ and (b) $\boldsymbol{d}_R=16$; baseline models using LAR-based inputs computed using trajectories observed within (c) $[t_0, t_0 + \Delta t]$ and (d) $[t_0, t_0 + 3\Delta t]$. Since LAR are frequency counts of GPS records which can be noisy, increasing the observation time interval is a proxy-method to increase the signal-to-noise ratio (SNR) of the LAR-based channels. Increasing $\boldsymbol{d}_R$ allows the contractive autoencoder, $\mathcal{F}_R$, to retain more semantically meaningful information that may be helpful for the reconstruction task. Since larger $\boldsymbol{d}_R$ results in lowering the maximum possible batch-size (governed by GPU memory) for training models, we refrain from increasing $\boldsymbol{d}_R$ beyond 16 given the practical constraints of heterogeneous compute clusters having GPUs with varied memory.

\subsection{Impact of Reachability Embeddings} 
Table \ref{results_table} shows the quantitative comparison of the performance for all 5 downstream tasks, quantified using AUPRC on the test set after model training converges (100 epochs), between the two variants each of baseline and reachability-based models. Figure \ref{prcurves} shows the corresponding precision-recall curves. Three key observations emerge: (i) increasing the observation interval (increasing SNR) for computing LAR and increasing $\boldsymbol{d}_R$ increases AUPRC --- for most values of recall, precision increases due to reduction in false positives; (ii) reachability-based models outperform the baseline models, including those using LAR computed by observing 3 times more trajectories: the AUPRC gain by the inferior reachability-based model ($\boldsymbol{d}_R = 8$) over the superior baseline model (observation interval $3\Delta t$) varies from 1.6\% for stop signs detection task to 18.4\% for the access point detection task; (iii) for the same observation time interval, $\Delta t$, simply replacing the LAR-based inputs by reachability embeddings ($\boldsymbol{d}_R = 16$) results in a AUPRC gain that varies from 4.1\% for the stop signs detection task to 23.3\% for the access point detection task. These observations conclusively demonstrate that reachability embeddings are more informative, denser representations of trajectory data requiring lesser trajectories (upto 67\% less) to compute. Thus, reachability embeddings  may be used to compute semantically meaningful representations of trajectories in geographical areas with less traffic or to build computer vision-based models for low-resource geospatial tasks.

\subsection{Visual Evaluation} 
\begin{figure}
    \begin{subfigure}{0.17\textwidth}
        \centering
        \includegraphics[width=\textwidth]{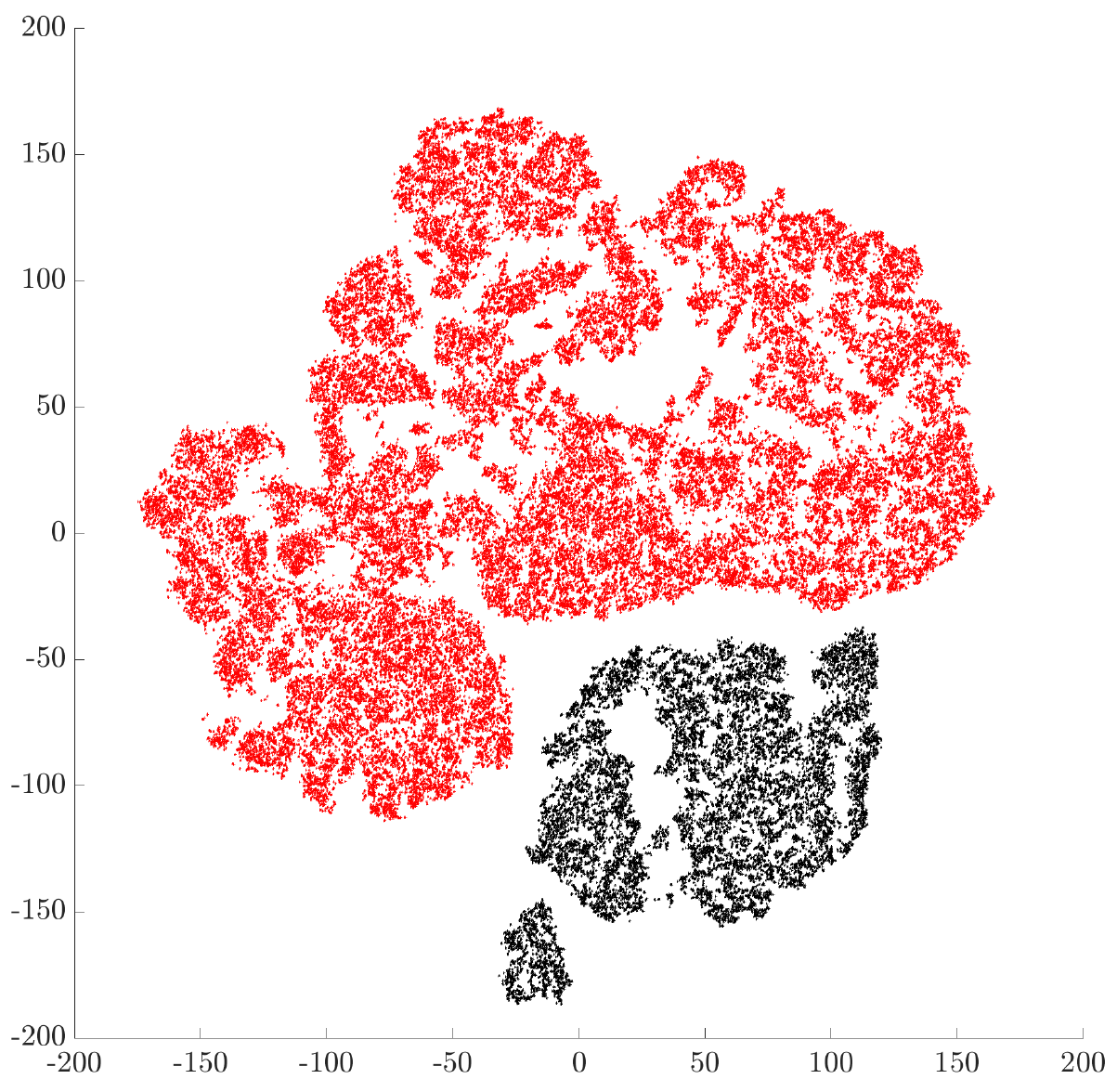}
        \caption{}
        \label{umap_reachability_embeddings}
    \end{subfigure}
    \hfill
    \begin{subfigure}{0.29\textwidth}
        \centering
        \includegraphics[width=\textwidth]{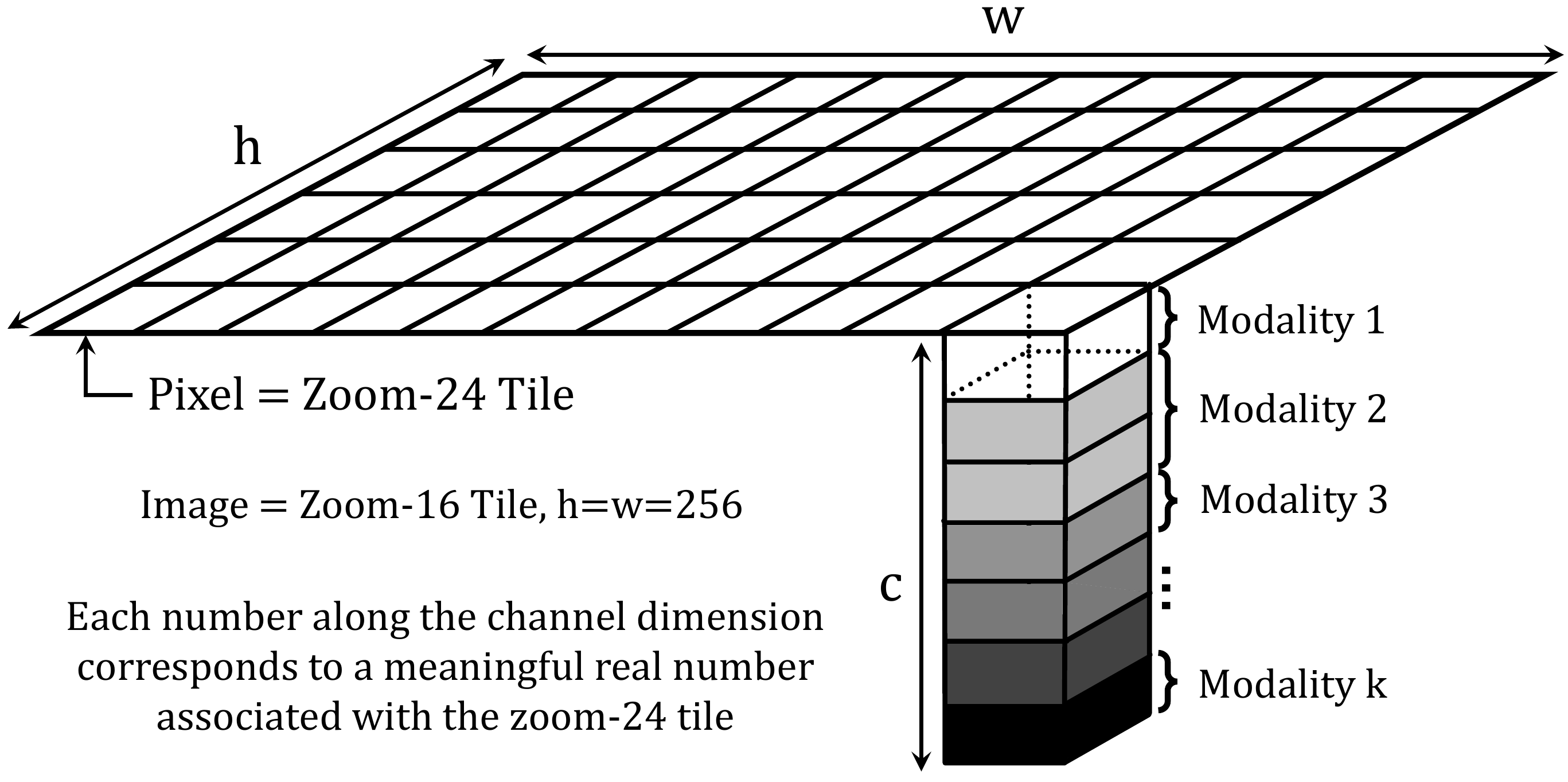}
        \caption{}
        \label{repn_algn_fusn}
    \end{subfigure}
    \caption{(a) UMAP of embeddings in a geographic location: highways nodes (black), otherwise (red). (b) Alignment and fusion of data modalities.}
\end{figure}

\begin{figure}
    \centering
    \begin{subfigure}[b]{0.24\textwidth}
         \centering
         \includegraphics[width=\textwidth]{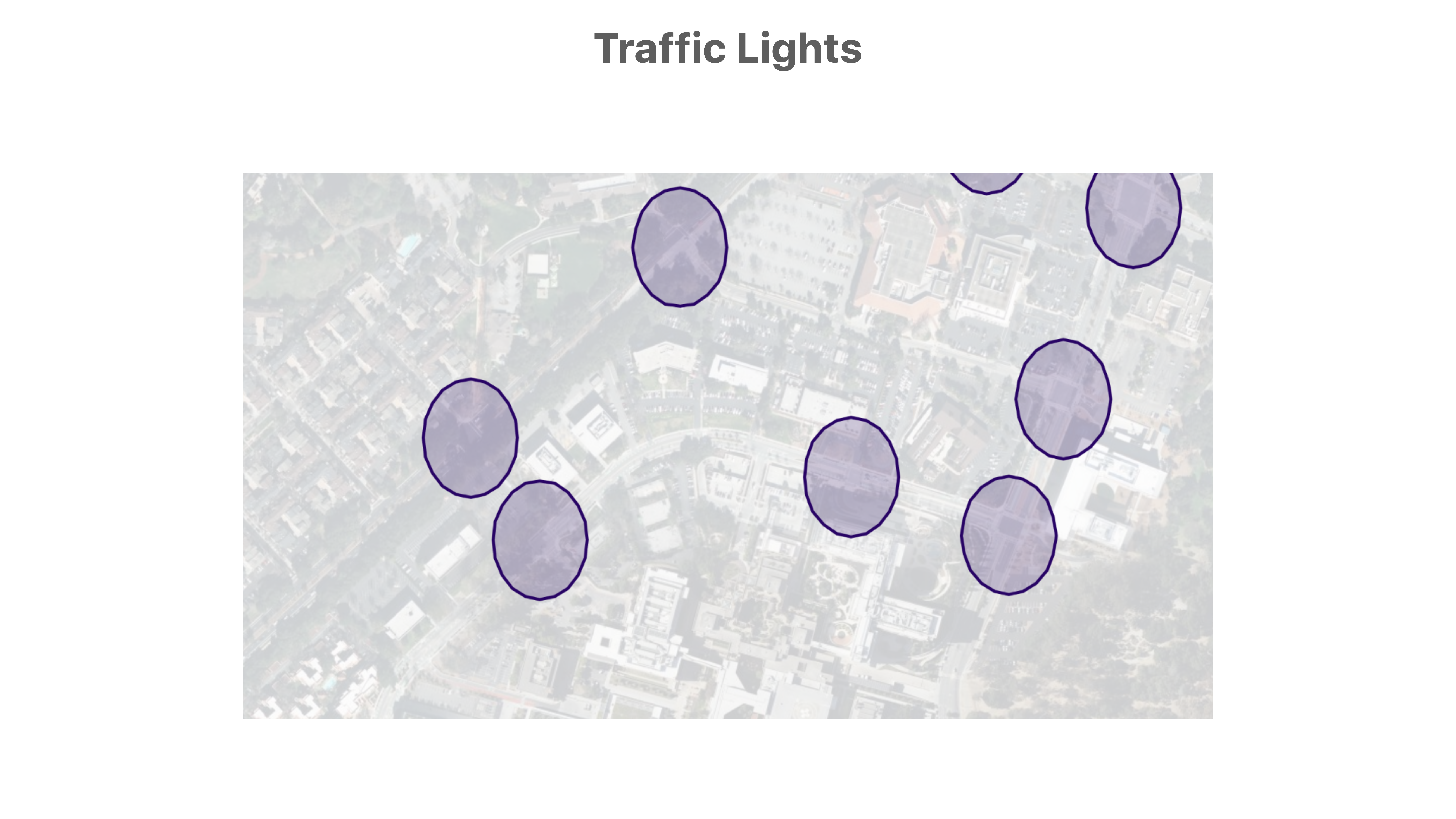}
         \caption{}
         \label{labels_tl}
     \end{subfigure}
     \hfill
     \begin{subfigure}[b]{0.24\textwidth}
         \centering
         \includegraphics[width=\textwidth]{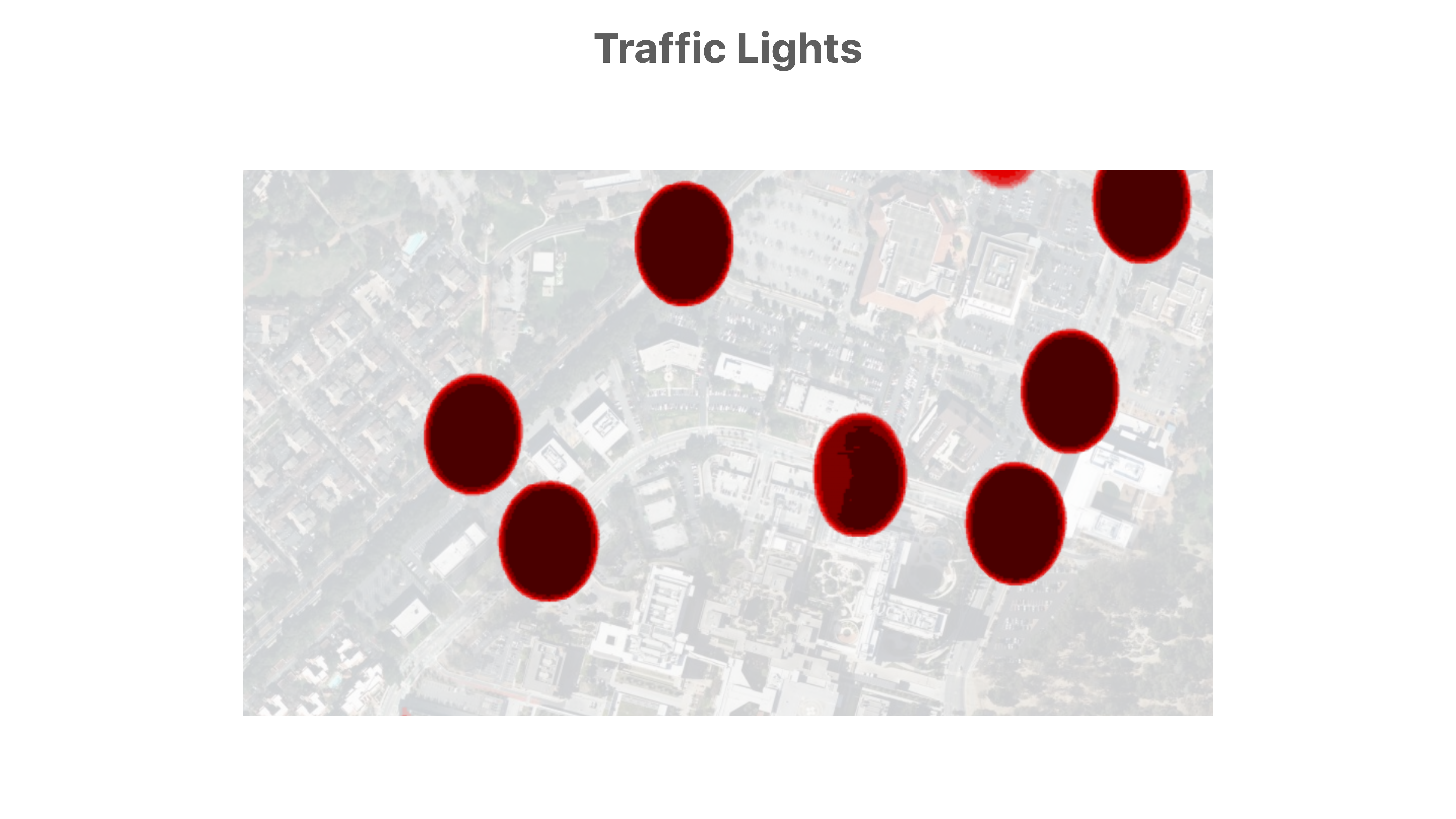}
         \caption{}
         \label{predictions_tl}
     \end{subfigure}
     
     \hfill
     
     \begin{subfigure}[b]{0.24\textwidth}
         \centering
         \includegraphics[width=\textwidth]{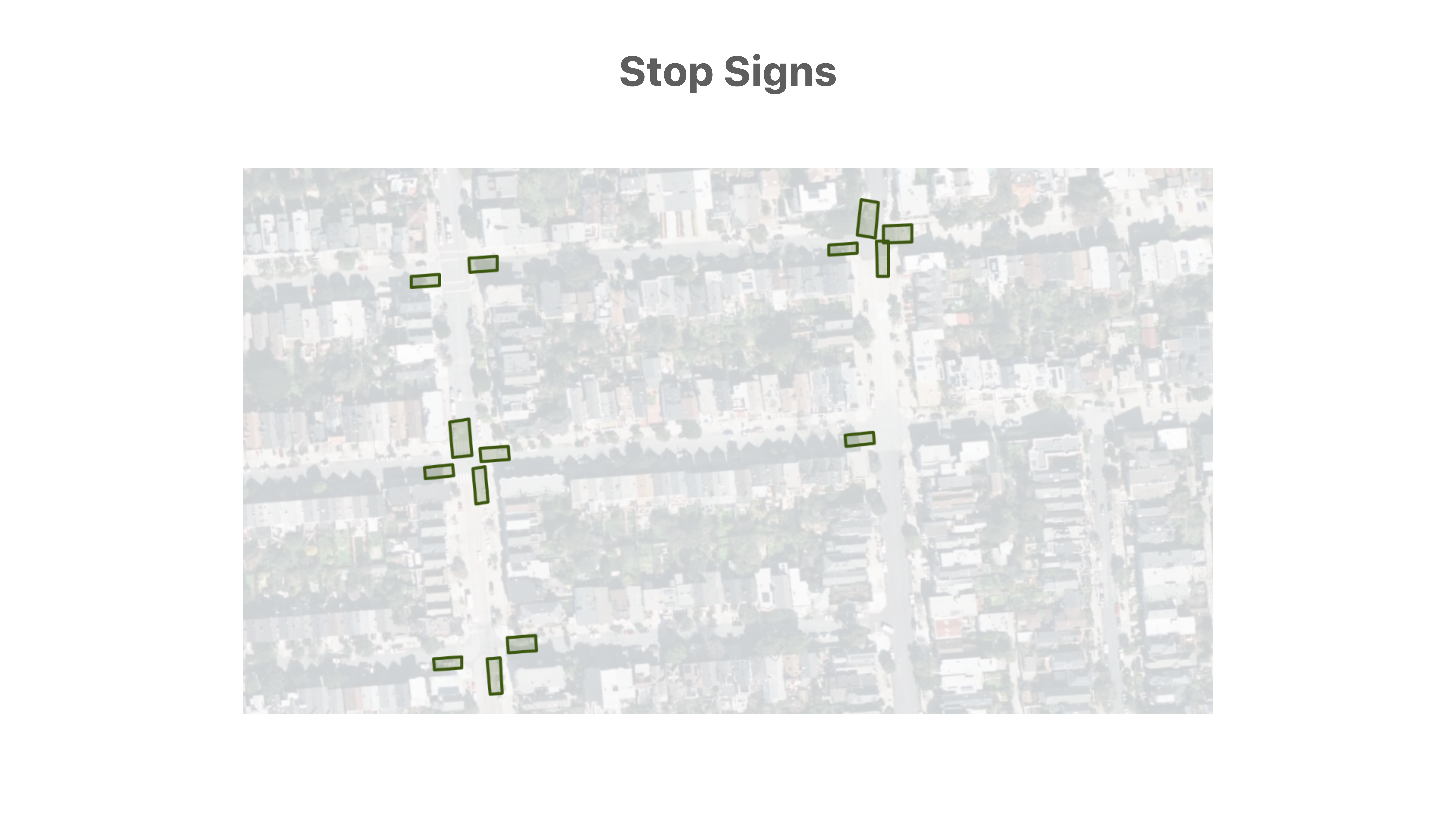}
         \caption{}
         \label{labels_ss}
     \end{subfigure}
     \hfill
     \begin{subfigure}[b]{0.24\textwidth}
         \centering
         \includegraphics[width=\textwidth]{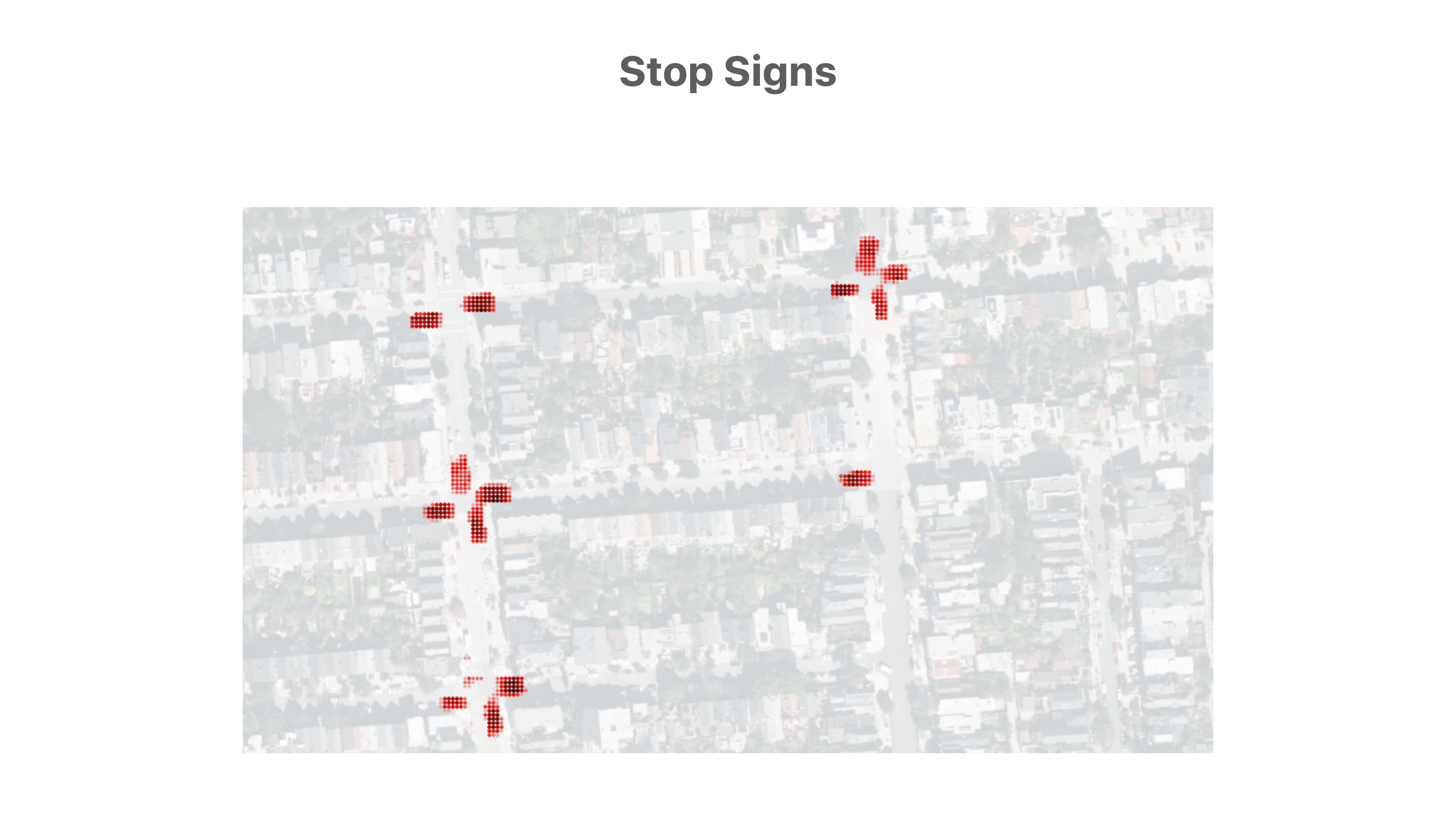}
         \caption{}
         \label{predictions_ss}
     \end{subfigure}
     
     \hfill
     
     \begin{subfigure}[b]{0.24\textwidth}
         \centering
         \includegraphics[width=\textwidth]{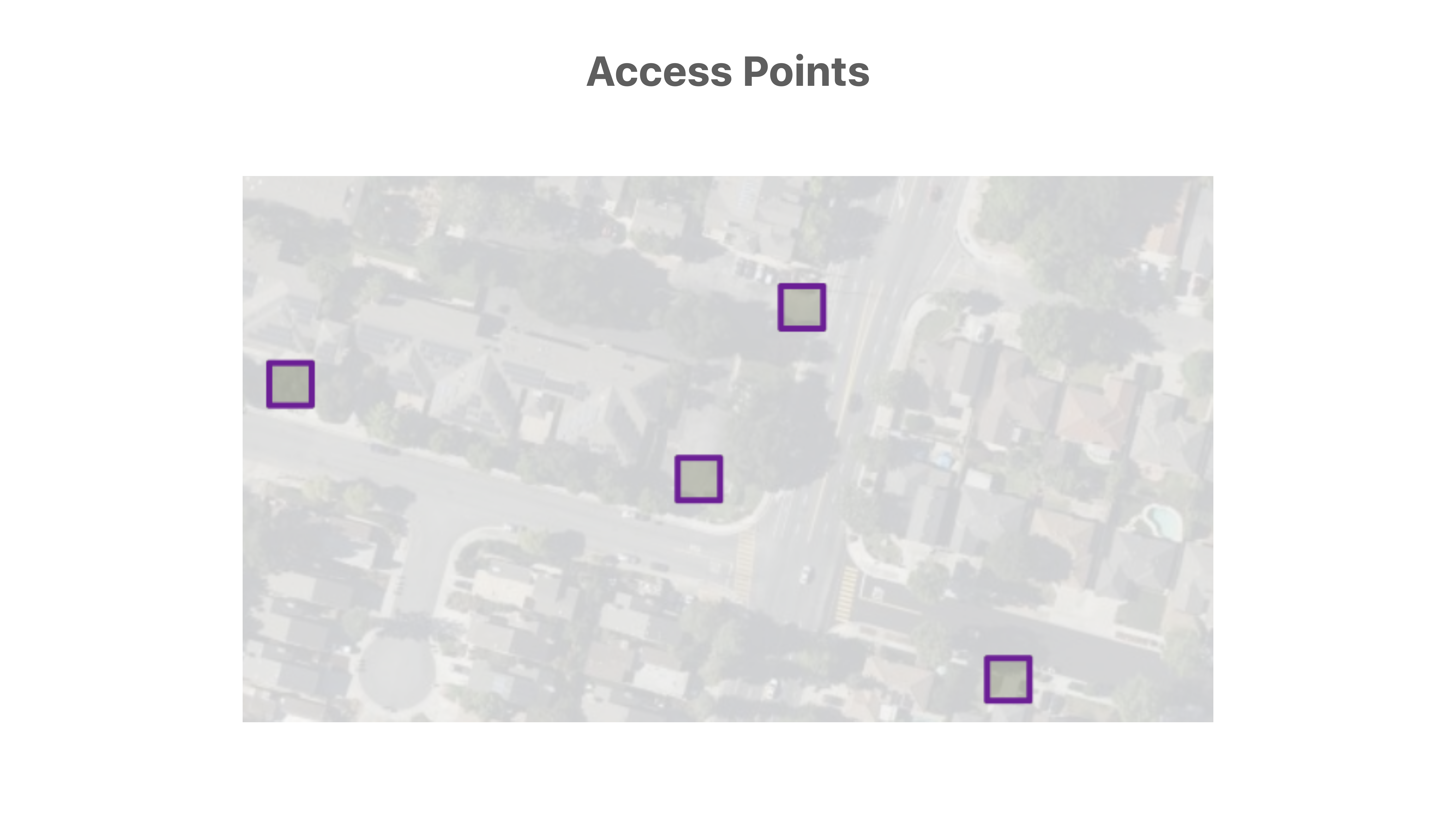}
         \caption{}
         \label{labels_draps}
     \end{subfigure}
     \hfill
     \begin{subfigure}[b]{0.24\textwidth}
         \centering
         \includegraphics[width=\textwidth]{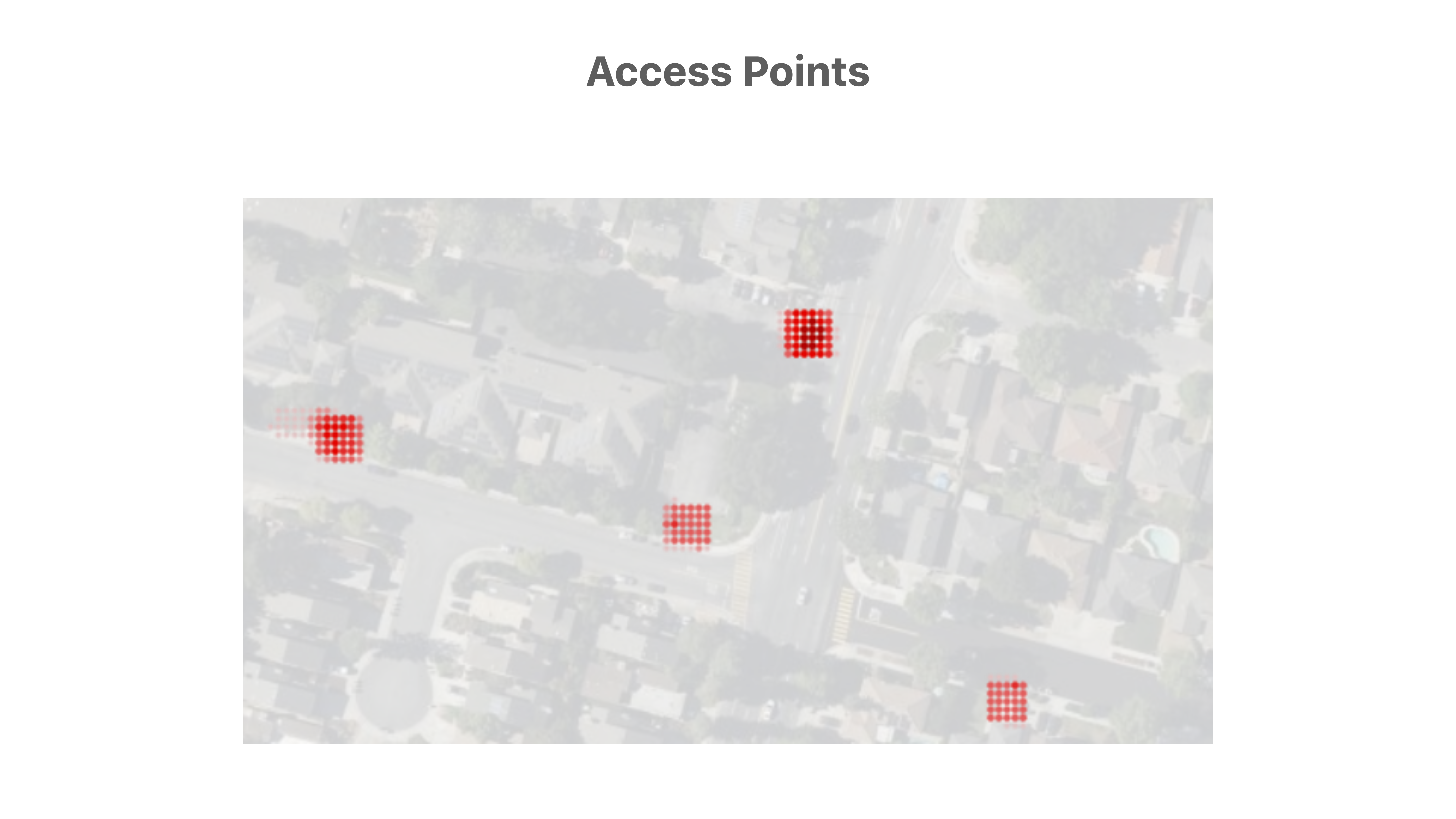}
         \caption{}
         \label{predictions_draps}
     \end{subfigure}
     \caption{Examples of labels and semantic segmentation model predictions from reachability-based models on the test set for 3 downstream tasks. Subfigures (a) and (b) show labels and predictions for traffic lights, (c) and (d) show labels and predictions for stop signs, and (e) and (f) show labels and predictions for access points. Satellite imagery is superposed for visual reference only.}
     \label{predictions_labels_3_downstream_tasks}
\end{figure}

Good learned representations ensure that similar entities have similar representations \cite{bengio13}. Since reachability embeddings are compressed representations of connectivity patterns, geographic locations with similar traffic patterns must have similar reachability embeddings. A visual demonstration of the semantics captured by reachability embeddings is shown by the UMAP projection of embeddings computed in a large geographical area in Figure \ref{umap_reachability_embeddings}. Black points denoting nodes on highways (mostly straight roads with few connections to neighbors) are clustered and cleanly separated from red points that denote residential roads. Visualizations of labels and predictions from reachability-based semantic segmentation models for three of the tasks in table \ref{results_table} is shown in Figure \ref{predictions_labels_3_downstream_tasks}.

\subsection{Multimodal Modeling with Reachability Embeddings} 
Representation, alignment, and fusion are three key challenges in multimodal machine learning \cite{baltruvsaitis2018}. The main advantage of reachability embeddings is the ability to combine mobility, satellite imagery, and imagery-like representations of graph data (e.g., RNP) for building multimodal models that either complement each other (when all data modalities are present) or supplement the incompleteness or corruption in a given modality. We choose the overpass, crosswalk, and access point detection tasks for this demonstration. In satellite imagery, tree canopies and buildings, among others, often occlude crosswalks and access points while some overpasses are hard to distinguish from junctions. We show that having mobility data can help the model achieve better performance. We compare the AUPRC of models using (i) only reachability embeddings, (ii) only satellite imagery, (iii) using satellite imagery and RNP, and (iv) using satellite imagery, RNP, and reachability embeddings. The design of reachability embeddings implies that aligning modalities at the pixel-level results in spatial alignment since pixels are zoom-24 tiles. The data modalities are concatenated along the channel dimension (called early fusion, shown in figure \ref{repn_algn_fusn}). Two neural architectures, UNet \cite{ronneberger2015} and SegNet \cite{badrinarayanan2017}, are compared for semantic segmentation. As can be seen in table \ref{multimodal_results_table}, multimodal inputs combining mobility data, satellite imagery, and road network graph improves model performance by 2--4\% compared to unimodal inputs, irrespective of neural architecture, which is facilitated by reachability embeddings.

\setlength{\belowcaptionskip}{-2pt}
\setlength{\textfloatsep}{5pt}
\begin{table}[hbt!]
    \scriptsize
    \caption{Reachability embeddings (RE, $\boldsymbol{d}_{R}=16$) facilitate multimodal modeling improving model performance (AUPRC, $\uparrow$ is better) for overpass (Task 1), crosswalk (Task 2), and access point (Task 3) detection. \label{multimodal_results_table}}
    \centering
    \begin{tabular}{P S S S S S S}
        \toprule
                           & \multicolumn{3}{c}{UNet \cite{ronneberger2015}}     & \multicolumn{3}{c}{SegNet \cite{badrinarayanan2017}} \\
        \cmidrule(lr){2-4}\cmidrule(lr){5-7}
        Input Modalities   & Task 1          & Task 2          & Task 3          & Task 1          & Task 2          & Task 3           \\
        \midrule
        RE                 & 0.925           & 0.976           & 0.864           & 0.917           & 0.981           & 0.881            \\
        Sat. Img.          & 0.923           & 0.968           & 0.856           & 0.901           & 0.970           & 0.874            \\
        Sat. Img., RNP     & 0.925           & 0.972           & 0.856           & 0.905           & 0.974           & 0.875            \\
        Sat. Img., RNP, RE & \textbf{0.949}  & \textbf{0.988}  & \textbf{0.895}  & \textbf{0.939}  & \textbf{0.992}  & \textbf{0.905}   \\
        \bottomrule
    \end{tabular}
\end{table}

\section{Related Work}\label{related_work}
\subsection{Graph Node Embeddings} 
While a number of deep learning methods have been proposed for generating node embeddings of a known graph \cite{cai18}, they can be grouped into three distinct categories: (a) techniques that efficiently sample random walk paths from a given graph and generate node embeddings by processing the paths as sequences using NLP-inspired methods like skip-gram \cite{mikolov13} such as \cite{perozzi14,grover16,dong17} or explicitly using recurrent neural networks (RNN) based on LSTMs or GRUs such as \cite{liu17}; (b) techniques that generate node embeddings by compressing matrix representations derived from graphs (e.g., adjacency matrix, positive point-wise mutual information matrix) such as \cite{cao16,wang16}; (c) techniques that generate node embeddings directly from the graph using graph neural networks (GNNs) and its variants \cite{zhou18} (e.g., graph convolutional network (GCN)). In contrast to techniques in category (a), (b), and (c), reachability embeddings \textit{infer} an implicit graphical structure for the earth's surface (earth surface graph) by treating geographic locations as nodes and observed GPS trajectories as allowed Markovian paths on this graph. Edges of the earth surface graph can be inferred using the observed paths. Reachability summaries are image-like representations for nodes deduced from observed paths and embedding generation is posed as an image reconstruction task using convolutional neural architectures as opposed to NLP-inspired models and RNN-based architectures used in techniques in category (a). While reachability summaries are matrix representations, they stand in contrast to techniques in category (b) since summaries are derived from observed trajectories and not from the graph or random walk paths sampled from a graph.

\subsection{Self-Supervised Learning for Computer Vision} 
SSL has been successfully applied to learn effective visual representations \cite{jing20,liu2020self}. Generative methods learn representations by modeling the data distribution \cite{donahue2019large,goodfellow2014gan,kingma2013vae} or reconstructing the input or feature \cite{gidaris20,vincent08}. Examples of heuristic, discriminative pretext tasks used to learn representations include context prediction \cite{doersch15}, solving jigsaw puzzles of image patches \cite{noroozi16}, and predicting image rotations \cite{gidaris18}. Among discriminative methods, contrastive methods currently achieve state-of-the-art performance in SSL \cite{chen2020simclr,grill20}. Reachability summary generation and its contractive reconstruction can together be viewed as the generative pretext task that encodes the co-occurrence relationships inherent in geospatial transitions (resulting from interaction of traffic and local transport infrastructure) to obtain reachability embeddings. 

\subsection{GPS Trajectory Embeddings} 
Most existing methods using GPS record or trajectory representations process trajectories similar to sequences with NLP-inspired methods like skip-gram or RNNs e.g., location similarity prediction \cite{crivellari19}, motion modality classification \cite{dabiri19,jiang17,gao17}, demographic attribute prediction \cite{solomon18}, and living pattern recognition \cite{cao19}. To the best of our knowledge, no prior work uses the Markovian concept of reachability and a computer vision-based SSL pretext task to learn self-supervised, contextual representations of geographic locations.

\section{Conclusions}\label{conclusions_and_future_work}
In this paper, we propose reachability embeddings, a novel, computer vision-based, self-supervised method to learn representations of geographic locations (zoom-24 tiles or nodes of the ESG) from observed GPS trajectories modeled as Markovian paths. Reachability summaries for each node are image-like representations that capture the inferred connectivity pattern based on evidence from observed trajectories (mobility data) of the inflow (emission transitions) and outflow (absorption transitions) of traffic at the node. Summaries are compressed to a vector representation for each node, called reachability embeddings, using the encoder of a contractive, fully-convolutional autoencoder trained to reconstruct reachability summaries. Reachability summary generation and its contractive reconstruction can together be viewed as the pretext task to obtain reachability embeddings. A theoretical interpretation of reachability embeddings is provided using the Chapman-Kolmogorov equations. The contractive regularization incentivizes robustness and invariance of embeddings to small perturbations in the summary leading embeddings of similar locations being close to each other in the learned, low-dimensional manifold. The proposed scalable, distributed algorithm (Algorithm \ref{algorithm_reachability_summary}) to generate reachability summaries, encoding spatial connectivity along with distance travelled and time taken during node transitions, shows good strong-scaling performance. Experiments in Section \ref{experiments_and_results} confirm that reachability embeddings are more informative and denser (uses up to 67\% less trajectory data compared to LAR) representations of geographic locations derived entirely from trajectory data leading to gains of 4--23\% in AUPRC on five different downstream supervised prediction tasks. Reachability embeddings are demonstrated to facilitate multimodal learning in geospatial computer vision with spatiotemporal mobility data as one of the data modalities. 

Comparing versions of embeddings computed by varying $t_0$, $\Delta t$, or both can be used for detecting changes in geospatial features over time, identifying locations with timed turn restrictions or road closures, identifying seasonal patterns, etc. As evidenced in this work, self-supervision is a promising approach to analyze spatiotemporal trajectory datasets.

\bibliographystyle{IEEEtran}
\bibliography{Paper_IEEEMDM2022}

\appendix

\subsection{Distributed, data-parallel implementation of Algorithm 1}\label{spark_implementation_of_algorithm_1}
Algorithm \ref{algorithm_reachability_summary_spark_implementation} presents the pseudo-code of a distributed, data-parallel implementation of Algorithm \ref{algorithm_reachability_summary} written using the Scala Dataset API in Apache Spark\footnote{https://spark.apache.org/} syntax. Apache Spark is an open-source, fault-tolerant, distributed, and data parallel cluster computing framework with support for distributed in-memory computation. Adopting an object-oriented style, Algorithm \ref{algorithm_reachability_summary_spark_implementation} uses the \texttt{Pair} and \texttt{IdxPair} case classes defined as follows. A transition between two nodes is represented by the \texttt{Pair} object that has four parameters: (i) \texttt{node1} and \texttt{node2} denote the nodes involved; (ii) \texttt{flg} denotes the type of transition from the vantage point of \texttt{node1} and is set to \texttt{flg=`a'} for the transition \texttt{node1}$\rightarrow$\texttt{node2} while it is set to \texttt{flg=`e'} for the \texttt{node2}$\rightarrow$\texttt{node1} transition; (iii) \texttt{cnt}, initialized to \texttt{1.0}, that counts the number of occurrences of the transition. An \texttt{IdxPair} object also represents a transition between two nodes, $\boldsymbol{s}, \boldsymbol{s'} \in \overline{\boldsymbol{V}}_{ES}$, and has four parameters: (i) \texttt{node} is one of $\boldsymbol{s}$ or $\boldsymbol{s'}$; (ii) \texttt{rmIdx} is the row-major index of $\boldsymbol{s'}$ in $\boldsymbol{\Psi}_{s}^{a}$ if \texttt{node}$=\boldsymbol{s}$ while it is the row-major index of $\boldsymbol{s}$ in $\boldsymbol{\Psi}_{s'}^{e}$ if \texttt{node}$=\boldsymbol{s'}$; (iii) \texttt{flg} is a flag set to \texttt{`a'} if \texttt{node}$=\boldsymbol{s}$ and to \texttt{`e'} if \texttt{node}$=\boldsymbol{s'}$, (iv) \texttt{cnt} is the frequency of the observed transition.

\begin{algorithm}
    \small
    \caption{Distributed, data-parallel implementation of Algorithm 1 for generating $\boldsymbol{\Psi}_{s}^{ea}\,\,\forall\,\,\boldsymbol{s} \in \overline{\boldsymbol{V}}_{ES}$}\label{algorithm_reachability_summary_spark_implementation}
    \begin{algorithmic}[1]
        \INPUT $\mathcal{T}_{t_0}^{t_0 + \Delta t} = (\boldsymbol{T}_1 \ldots \boldsymbol{T}_M)$, $|\boldsymbol{T}_i| = n_i \,\,\forall\,\, i \in [1,M]$, $\boldsymbol{T}_i = (\boldsymbol{p}_1^i, \ldots, \boldsymbol{p}_{n_i}^i)$ where $\boldsymbol{p}_k^i = (\boldsymbol{z}_k^i, \boldsymbol{t}_k^i)$
        \OUTPUT $\boldsymbol{\Psi}_s^e, \boldsymbol{\Psi}_s^a \,\,\forall\,\, s \in \overline{\boldsymbol{V}}_{ES}$
        \PARAMETERS $\boldsymbol{\delta}_R$
        \INITIALIZE Ingest $\mathcal{T}_{t_0}^{t_0 + \Delta t}$ as \texttt{trj}
        \Procedure{ReachabilitySummaryGenerator}{$\phantom{}$}
            \State \texttt{result} = \texttt{trj}.\texttt{flatMap}(\texttt{t} $\Rightarrow$ \texttt{createAPairs}(\texttt{t}))
            \State .\texttt{flatMap}(\texttt{p} $\Rightarrow$ \texttt{createEPairs}(\texttt{p}))
            \State .\texttt{map}(\texttt{p} $\Rightarrow$ \texttt{getIdxForDestinationTile}(\texttt{p}))
            \State .\texttt{map}(\texttt{p} $\Rightarrow$ (\texttt{p.cTile},\texttt{p.dIdx},\texttt{p.flg}) $\rightarrow$ \texttt{p.cnt})
            \State .\texttt{reduceByKey}((\texttt{v1}, \texttt{v2}) $\Rightarrow$ \texttt{v1} + \texttt{v2})
            \State .\texttt{map}((\texttt{k},\texttt{v}) $\Rightarrow$ \texttt{k.cTile} $\rightarrow$ (\texttt{k.dIdx},\texttt{k.flg},\texttt{v}))
            \State .\texttt{groupByKey}()
            \State .\texttt{map}((\texttt{k},\texttt{v}) $\Rightarrow$ Create $\boldsymbol{\Psi}_{k}^{e},\boldsymbol{\Psi}_{k}^{a},\boldsymbol{\Psi}_{k}^{ea}$ for tile \texttt{k})
            \State .\texttt{saveToDistributedStorage}($\boldsymbol{\Psi}_{k}^{ea}$ $\forall$ $k \in \overline{\boldsymbol{V}}_{ES}$)
        \EndProcedure
        \Function{createAPairs}{$\boldsymbol{T}_i$}
            \For{$k \leftarrow 1$ to $n_i$}
                \For{$l \leftarrow k$ to $n_i$}
                    \If{$\boldsymbol{z}_l^i \in \boldsymbol{N}_{z_k^i}^{{\delta_R}}$}
                        \State \Return \texttt{\textbf{Pair}(}$\boldsymbol{z}_k^i,\boldsymbol{z}_l^i,$\texttt{`a',1.0)}
                    \EndIf
                \EndFor
            \EndFor
        \EndFunction
        \Function{createEPairs}{\texttt{p: \textbf{Pair}}}
            \State \Return \texttt{(p,\textbf{Pair}(p.node2,p.node1,`e',p.cnt))}
        \EndFunction
        \Function{getIdxForDestinationTile}{\texttt{p:\textbf{Pair}}}
            \State $\boldsymbol{s}, \boldsymbol{s}' \leftarrow$ \texttt{p.node1,p.node2}
            \State \texttt{rmIdx} $\leftarrow \boldsymbol{L}(\boldsymbol{x}_{s'}-\boldsymbol{x}_{s} + \boldsymbol{\delta}_R) + (\boldsymbol{y}_{s'}-\boldsymbol{y}_{s} + \boldsymbol{\delta}_R)$
            \State \Return \texttt{\textbf{IdxPair}(p.node1,rmIdx,p.flg,p.cnt)}
        \EndFunction
    \end{algorithmic}
\end{algorithm}

Line 2 of Algorithm \ref{algorithm_reachability_summary_spark_implementation} creates valid absorption transitions as \texttt{Pair} objects from $\boldsymbol{T}_i$. Line 3 computes the corresponding emission transition for every absorption transition. All \texttt{Pair} objects representing emission and absorption transitions are converted to corresponding \texttt{IdxPair} objects in line 4. Lines 5 and 6 count the observed frequency of both absorption and emission transitions between all pairs of nodes in $\overline{\boldsymbol{V}}_{ES}$ from all trajectories in $\boldsymbol{\mathcal{T}}(t_0,\Delta t)$. All associated transitions are gathered in line 8 to compute the absorption and emission channels which are concatenated in line 9 to obtain the reachability summaries for all nodes in $\overline{\boldsymbol{V}}_{ES}$. Line 10 saves the output into distributed storage.

\subsection[Modified analyzeTrajectories procedure]{Modified $\mathtt{analyzeTrajectories}$ procedure}\label{appendix:createAPairs}
Algorithm \ref{algorithm_modified_createEPairs} presents the modified version of {\small \texttt{analyzeTrajectories}} procedure when distance and time weighting, discussed in Section \ref{encoding_more_information_in_reachability_summaries}, are used. Note that the same transition, $(\boldsymbol{s},\boldsymbol{s}')$, can have different counts within a single trajectory if the transitions take two different possible paths during which the distance covered and time taken to cover the distance are different. Line 5 in Algorithm \ref{algorithm_modified_createEPairs} uses the haversine formula \cite{wikihaversine20} to calculate the great-circle distance between the centroids of two zoom-24 tiles. One can analogize the weight decays for nodes $\boldsymbol{s}' \in \boldsymbol{N}_{s}^{\delta_R}$ in the reachable neighborhood of node $\boldsymbol{s}$ to the soft-attention mechanism in NLP where the magnitude of the weight decay can be thought of as the ``attention" payed by node $\boldsymbol{s}$ to the node $\boldsymbol{s}'$ in the calculation of the reachability summary, $\boldsymbol{\Psi}_s^{ea}$.

\begin{algorithm}
    \small
    \caption{Encode distance \& time weighting in $\boldsymbol{\Psi}_{s}^{ea}$}\label{algorithm_modified_createEPairs}
    \begin{algorithmic}[1]
        \INPUT $\boldsymbol{T}_i \in \mathcal{T}_{t_0}^{t_0 + \Delta t}$, $|\boldsymbol{T}_i| = n_i$, $\boldsymbol{T}_i = (\boldsymbol{p}_1^i, \ldots, \boldsymbol{p}_{n_i}^i)$ where $\boldsymbol{p}_k^i = (\boldsymbol{z}_k^i,\boldsymbol{t}_k^i)$
        \OUTPUT Map $\boldsymbol{S}$
        \PARAMETERS $\boldsymbol{\delta}_R, \sigma_d, \sigma_t$
        \INITIALIZE $\boldsymbol{T}_i = (\overline{\boldsymbol{p}}_1^i, \ldots, \overline{\boldsymbol{p}}_{n_i}^i)$ such that the 3-tuple $\overline{\boldsymbol{p}}_k^i = (\boldsymbol{z}_k^i,\boldsymbol{t}_k^i, \boldsymbol{d}_k^i=0)\,\,\forall\,\,\boldsymbol{T}_i \in \mathcal{T}_{t_0}^{t_0 + \Delta t}$. Let $\overline{n} = \max\{n_1, \ldots, n_M\}$
        \Function{analyzeTrajectories}{$\mathcal{T}_{t_0}^{t_0 + \Delta t}$}\Comment{$\mathcal{O}(M\overline{n}^2)$}
            \State Initialize map $\boldsymbol{S} = \emptyset$\
            \For{trajectory $\boldsymbol{T}_i \in \mathcal{T}_{t_0}^{t_0 + \Delta t}$}
                \For{$k \leftarrow 2$ to $n_i$}\Comment{$\mathcal{O}(n_i)$}
                    \State $\Delta d$ = haversineDistance$(\boldsymbol{z}_{(k-1)}^i,\boldsymbol{z}_k^i)$
                    \State $\boldsymbol{d}_k^i \leftarrow \boldsymbol{d}_{(k-1)}^i + \Delta d$  
                \EndFor
                \For{$k \leftarrow 1$ to $n_i$}\Comment{$\mathcal{O}(n_i^2)$}
                    \For{$l \leftarrow k$ to $n_i$}
                        \If{$\boldsymbol{z}_l^i \in \boldsymbol{N}_{z_k^i}^{{\delta_R}}$}
                            \State if $\boldsymbol{z}_k^i \notin \boldsymbol{S}$, $\boldsymbol{S}[\boldsymbol{z}_k^i] = (\boldsymbol{S}_{z_k^i}^{e} = \emptyset, \boldsymbol{S}_{z_k^i}^{a} = \emptyset)$
                            \State if $\boldsymbol{z}_l^i \notin \boldsymbol{S}$, $\boldsymbol{S}[\boldsymbol{z}_l^i] = (\boldsymbol{S}_{z_l^i}^{e} = \emptyset, \boldsymbol{S}_{z_l^i}^{a} = \emptyset)$
                            \State $r_{z_k^i}(z_l^i) \leftarrow \texttt{getIndex}(\boldsymbol{z}_k^i, \boldsymbol{z}_l^i)$
                            \State $r_{z_l^i}(z_k^i) \leftarrow \texttt{getIndex}(\boldsymbol{z}_l^i, \boldsymbol{z}_k^i)$
                            \State $\Delta_{d,i}^{(z_k^i,z_l^i)} \leftarrow (\boldsymbol{d}_l^i-\boldsymbol{d}_k^i)$
                            \State $\Delta_{t,i}^{(z_k^i,z_l^i)} \leftarrow (\boldsymbol{t}_l^i-\boldsymbol{t}_k^i)$
                            \State $\Delta\boldsymbol{w}(\boldsymbol{z}_k^i,\boldsymbol{z}_l^i;\sigma_d,\sigma_t) \leftarrow$  Section (\ref{encoding_more_information_in_reachability_summaries})
                            \State $c_{z_k^i}^{a}(z_l^i) = \Delta\boldsymbol{w}(\boldsymbol{z}_k^i,\boldsymbol{z}_l^i;\sigma_d,\sigma_t)$
                            \State $c_{z_l^i}^{e}(z_k^i) = \Delta\boldsymbol{w}(\boldsymbol{z}_k^i,\boldsymbol{z}_l^i;\sigma_d,\sigma_t)$
                            \State $\boldsymbol{S}_{z_k^i}^{a}[r_{z_k^i}(z_l^i)] \pluseq c_{z_k^i}^{a}(z_l^i)$
                            \State $\boldsymbol{S}_{z_l^i}^{e}[r_{z_l^i}(z_k^i)] \pluseq c_{z_l^i}^{e}(z_k^i)$
                        \EndIf
                    \EndFor
                \EndFor
            \EndFor
            \State \Return $\boldsymbol{S}$
        \EndFunction
    \end{algorithmic}
\end{algorithm}

\subsection{T-Drive Dataset Pre-processing}\label{tdrive_preprocessing}
To demonstrate scalability of Algorithm \ref{algorithm_reachability_summary} in Section \ref{scalability_of_algorithm}, the publicly available T-Drive dataset \cite{msrtaxi11} from Microsoft Research, containing location coordinates (latitude-longitude pairs) and timestamps for 7 days of 10,357 taxis in Beijing, is used. To transform these into a trajectory dataset, $\boldsymbol{\mathcal{T}}(t_0, \Delta t)$, the data is pre-processed so that the daily chronological sequence of coordinate-timestamp pairs for each taxi is considered as one contiguous trajectory leading to $|\boldsymbol{\mathcal{T}}(t_0, \Delta t)| = M = 68,851$, where $t_0 = \text{February 02, 2008}$ and $\Delta t = \text{7 days}$. Coordinates are converted to zoom-24 tiles to match the format required for consumption by Algorithm \ref{algorithm_reachability_summary}. Three datasets of 2000, 8000, and 64000 trips are randomly sampled from this processed dataset for the scaling analysis in Section \ref{scalability_of_algorithm}.

\subsection{The Contractive Autoencoder}\label{contractive_autoencoder_architecture}
\begin{figure*}
    \centering
    \includegraphics[width=0.82\textwidth]{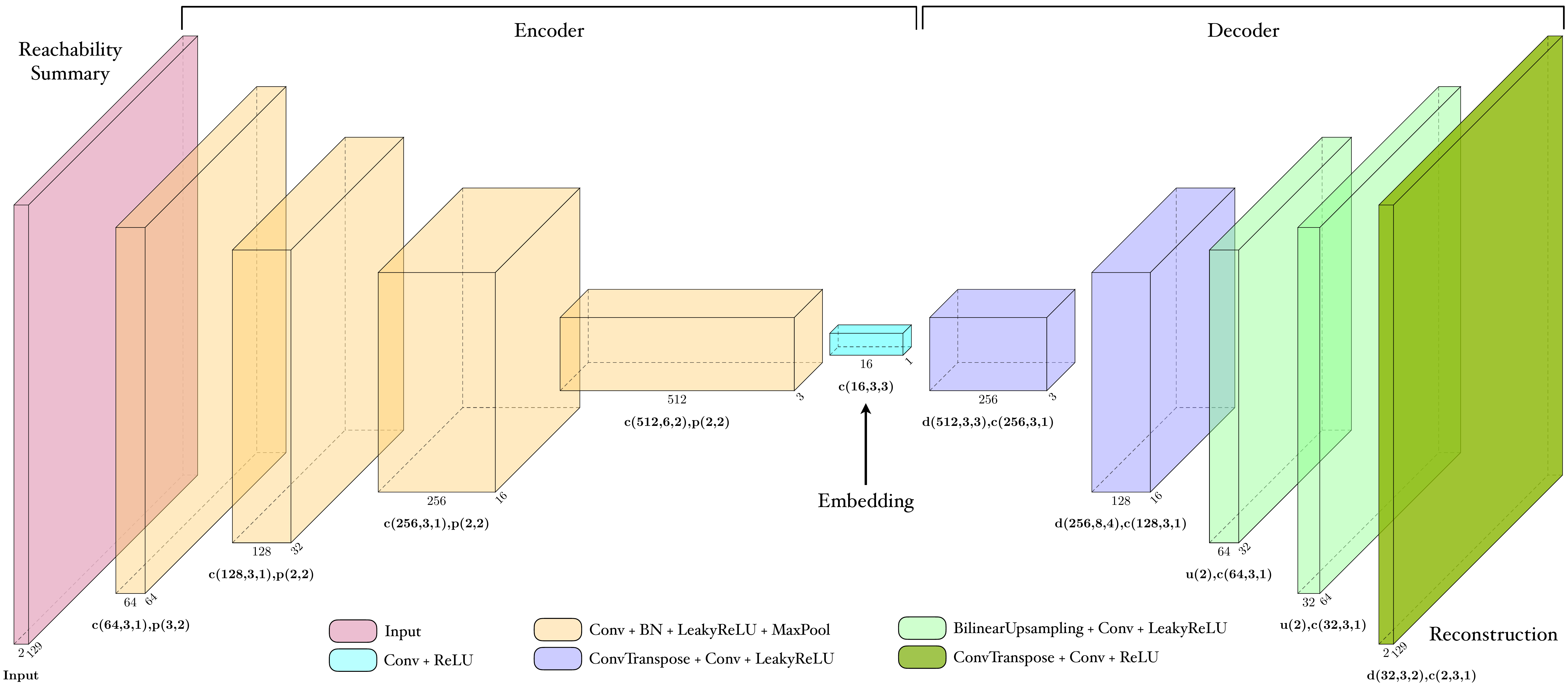}
    \caption{Contractive autoencoder architecture. Blocks describe outputs from preceding layer detailed in block caption. Layers denoted using $c(f, k,s)$, $p(k,s)$, $d(f, k,s)$, and $u(m)$, where, $c$ is convolution, $p$ is max-pooling, $d$ is transposed convolution, $u(m)$ is bilinear upsampling with multiplicity factor $m$, $f$ is number of filters, $k$ is kernel size, and $s$ is stride.}
    \label{contractive_autoencoder}
\end{figure*}

The neural architecture of the custom-designed contractive autoencoder, $\mathcal{F}_{R}$, is presented in Figure \ref{contractive_autoencoder}. The encoder, $\mathcal{F}_{R}^{e}$, has 5.2 million trainable parameters and the decoder, $\mathcal{F}_{R}^{d}$, has 5.8 million parameters, respectively. Using cross-validation, $\lambda$ in equation (\ref{total_loss}) is found to be 0.5. The autoencoder is implemented and trained using the Python API for TensorFlow. Distributed training of the model uses 8 NVIDIA Tesla V100 GPUs with a batch-size of 256 examples per GPU. For predicting reachability embeddings, distributed prediction on 120 NVIDIA Tesla V100 GPUs with a batch-size of 2000 examples per GPU is used.

\subsection{Interpreting Reachability Embeddings using the Chapman-Kolmogorov Equations}\label{methodology_chapman_kolmogorov}
The Markovian transitions in $\boldsymbol{V}_{ES}$ define a \textit{transition probability matrix} \cite{stirzaker92}, $\mathbf{P}_{ES} = [\mathbb{P}_{ES}^{(s,s')}]\in[0,1]^{|V_{ES}|\times|V_{ES}|}$,
\begin{equation*}\label{probability_calculation}
    \boldsymbol{Z} = \sum\limits_{s'' \in V_{ES}} \boldsymbol{w}_{ES}^{(s,s'')}\,\,;\,\,\mathbb{P}_{ES}^{(s,s')} = \boldsymbol{w}_{ES}^{(s,s')}/\boldsymbol{Z} \,\,\forall\,\,\boldsymbol{s}, \boldsymbol{s}' \in \boldsymbol{V}_{ES}.
\end{equation*}
Matrices, $\boldsymbol{\Psi}_{s}^{e}$ and $\boldsymbol{\Psi}_{s}^{a}$, defined in equations (\ref{emission_matrix}) and (\ref{absorption_matrix}), are closely related to $\mathbf{P}_{ES}$. For any square matrix, $X$, let $\text{sum}(X)$ denote the sum of all the matrix elements and $\text{scale}(X) = X/\text{sum}(X)$. Then, $\text{scale}(\Psi_s^e)$ and $\text{scale}(\Psi_s^a)$ are spatial proximity-preserving rearrangements in matrix form of the column and row vectors of $\mathbf{P}_{ES}$ corresponding to node $\boldsymbol{s}$, respectively. The Chapman-Kolmogorov equations (CKE) \cite{stirzaker92} are a fundamental identity of Markov chains and the inspiration for the construction of reachability summaries described in section \ref{methodology_reachability_embeddings}. While CKE is a more general result, in this paper we consider the specific case of \textit{discrete} and \textit{homogeneous} \cite{stirzaker92} Markov chains.  

Consider the state space $\boldsymbol{V}_{ES}$. All states are discrete and all Markovian transitions in the state space are modeled as homogeneous. For any node, $\boldsymbol{s}$, we wish to learn a representation of $\boldsymbol{s}$ that captures its contribution to connecting any two arbitrarily chosen nodes $\boldsymbol{s}',\boldsymbol{s}'' \in \boldsymbol{V}_{ES}$. Specifically, we wish to encode the information captured by the two-step transition probability from $\boldsymbol{s'}$ to $\boldsymbol{s}''$ passing through $\boldsymbol{s}$ into the representation for $\boldsymbol{s}$, for all choices of $\boldsymbol{s}', \boldsymbol{s}'' \in \boldsymbol{V}_{ES}$ as evidenced by transitions deduced from observed trajectories in the set $\mathcal{T}(t_0, \Delta t)$. This is in contrast to algorithms like \cite{dong17,grover16,perozzi14} which synthetically construct random walks on a graph. Setting $k=1, n=2$, for chosen state $\boldsymbol{s} \in \boldsymbol{V}_{ES}$ and $\forall\,\,\boldsymbol{s}',\boldsymbol{s}'' \in \boldsymbol{V}_{ES}$, it follows from equation CKE that
\begin{equation}\label{one_step_transitions}
    \left(\mathbf{P}_{ES}^{(s',s'')}\right)^{2} = \mathbf{P}_{ES}^{(s',s)} \mathbf{P}_{ES}^{(s,s'')} + \sum\limits_{z \in V_{ES} \setminus \{s\}} \mathbf{P}_{ES}^{(s',z)} \mathbf{P}_{ES}^{(z,s'')} 
\end{equation}
where, the first term on the right-hand side, henceforth denoted as $C_{s}^{(s',s'')}$, measures the contribution of state $\boldsymbol{s}$ to the total two-step transition probability from $\boldsymbol{s}'$ to $\boldsymbol{s}''$. Note that only the subset of trajectories in $\boldsymbol{\mathcal{T}}(t_0,\Delta t)$ that pass through $\boldsymbol{s}$, denoted as $\boldsymbol{\mathcal{T}}(t_0, \Delta t)(\boldsymbol{s})$, contribute to $C_{s}^{(s',s'')}$. Summing $C_{s}^{(s',s'')}$ over all possible $\boldsymbol{s}',\boldsymbol{s}'' \in \boldsymbol{V}_{ES}$, which is the total contribution of $\boldsymbol{s}$ to all transitions in $\boldsymbol{V}_{Es}$, denoted as $C_s$, and if only transitions from $\boldsymbol{N}_{s}^{\delta_R}$ are considered valid, we obtain  
\begin{align*}
    \begin{split}
        C_s = \sum_{s',s'' \in V_{ES}} C_{s}^{(s',s'')} &= \sum_{s' \in \Xi_{s}^{\delta_R}} \mathbf{P}_{ES}^{(s',s)} \sum_{s'' \in \Lambda_{s}^{\delta_R}} \mathbf{P}_{ES}^{(s,s'')}.
    \end{split}
\end{align*}
If the total number of Markovian transitions in $\boldsymbol{\mathcal{T}}(t_0, \Delta t)(\boldsymbol{s})$ that start at $\boldsymbol{s}$ is $\nu_s^a$, and that end at $\boldsymbol{s}$ is $\nu_s^e$, we have
\begin{align*}
    \begin{split}
        \nu_s^a \nu_s^e C_s = \sum_{s' \in \Xi_{s}^{\delta_R}} \left(\nu_s^e \mathbf{P}_{ES}^{(s',s)}\right) \sum_{s'' \in \Lambda_{s}^{\delta_R}} \left(\nu_s^a \mathbf{P}_{ES}^{(s,s'')}\right) = \boldsymbol{\psi}_s^{e \top} \boldsymbol{\psi}_s^a.
    \end{split}
\end{align*}
Here, $\boldsymbol{\psi}_s^e$ and $\boldsymbol{\psi}_s^a$ are $\boldsymbol{L}^2$-dimensional column vectors of the emission and absorption probabilities of $\boldsymbol{s}$ corresponding to the $\boldsymbol{L}^2$ states in $\boldsymbol{N}_{s}^{\delta_R}$. When $\boldsymbol{\psi}_s^e$ and $\boldsymbol{\psi}_s^a$ are reshaped to spatial-relation-preserving $\boldsymbol{L} \times \boldsymbol{L}$-dimensional matrices based on the relative positions of the nodes in $\boldsymbol{N}_{s}^{\delta_R}$ on $\boldsymbol{G}_{ES}$, they yield $\boldsymbol{\Psi}_s^e$ and $\boldsymbol{\Psi}_s^a$, respectively. Thus, $\boldsymbol{\Psi}_s^e$ and $\boldsymbol{\Psi}_s^a$ exactly capture the contribution of $\boldsymbol{s}$ to connecting nodes in $\boldsymbol{N}_{s}^{\delta_R}$\footnote{Analogous to the role layover airports or connection hubs play (e.g., Dubai for Emirates) in 1-stop flights. The key idea is to use the information of the set of flights flying \textit{into} the airport and the information of the set of flights flying \textit{out of} the airport to learn the representation of the layover airport.} where it acts as the intermediary, thereby capturing essential spatial connectivity information. The CKE (equation \ref{one_step_transitions}) are reinterpreted using the lens of reachability-based emission and absorption channels by noting
\begin{align}
    \begin{split}
        \sum_{s',s'' \in V_{ES}} \left(\mathbf{P}_{ES}^{(s',s'')}\right)^{2} = \sum\limits_{z \in V_{ES}} \boldsymbol{\psi}_z^{e \top} \boldsymbol{\psi}_z^a.
    \end{split}
\end{align}

\end{document}